\documentclass[]{alaya}
\usepackage{makecell}
\usepackage{wrapfig}
\usepackage{tabularx}
\usepackage{textcomp}
\usepackage{stfloats}
\usepackage{url}
\usepackage{verbatim}
\usepackage{titlesec}
\usepackage{tocloft}
\usepackage{adjustbox}
\usepackage{multirow}
\usepackage{pifont}
\usepackage[sc]{mathpazo}
\usepackage{tikz}
\usepackage{comment}
\usepackage{amsmath,amssymb}
\usepackage{colortbl}
\usepackage[numbers,sort&compress]{natbib}
\usepackage{color}
\usepackage{booktabs} 
\usepackage{hyperref}
\usepackage{graphicx}
\usepackage{subcaption}
\usepackage{multirow}
\usepackage{subcaption}
\RequirePackage{xspace}
\makeatletter
\DeclareRobustCommand\onedot{\futurelet\@let@token\@onedot}
\def\@onedot{\ifx\@let@token.\else.\null\fi\xspace}
\usepackage[most]{tcolorbox}
\usepackage{xcolor}
\usepackage{array}
\usepackage{tabularx}
\usepackage{siunitx} 
\usepackage{makecell}
\usepackage[table]{xcolor}
\usepackage{caption}
\definecolor{headerpurple}{HTML}{d8d2fc}
\definecolor{rowgray}{gray}{0.95}

\makeatother

\definecolor{adptorange}{RGB}{248, 205, 172}
\definecolor{cmpblue}{RGB}{189, 215, 238}
\definecolor{cmpblue}{RGB}{189, 215, 238}

\definecolor{our_red}{RGB}{232,157,160}
\definecolor{our_blue}{RGB}{136,206,230}
\definecolor{our_orange}{RGB}{246,200,168}
\definecolor{our_green}{RGB}{178,211,164}

\definecolor{attn_code0}{RGB}{247,215,200}
\definecolor{attn_code1}{RGB}{238,169,139}
\definecolor{mlp_code0}{RGB}{204,201,221}
\definecolor{mlp_code1}{RGB}{102,95,153}
\definecolor{mygray}{HTML}{f0f0f0}

\definecolor{token_blue}{RGB}{84, 120, 140}

\usepackage{pifont}
\usepackage{bbding}
\usepackage{fontawesome}
\usepackage{xspace}

\usepackage{float}

\newlength\savewidth\newcommand\shline{\noalign{\global\savewidth\arrayrulewidth \global\arrayrulewidth 1pt}\hline\noalign{\global\arrayrulewidth\savewidth}}
\newcommand{\tablestyle}[2]{\setlength{\tabcolsep}{#1}\renewcommand{\arraystretch}{#2}\centering\footnotesize}

\newcolumntype{x}[1]{>{\centering\arraybackslash}p{#1pt}}
\newcolumntype{y}[1]{>{\raggedright\arraybackslash}p{#1pt}}
\newcolumntype{z}[1]{>{\raggedleft\arraybackslash}p{#1pt}}

\renewcommand{\paragraph}[1]{\vspace{1mm}\noindent\textbf{#1}}
\usepackage{colortbl}
\usepackage{xcolor}
\usepackage{wrapfig}

\renewcommand{\paragraph}[1]{\vspace{1.25mm}\noindent\textbf{#1}}

\usepackage{algorithm}
\usepackage{listings}

\definecolor{codeblue}{rgb}{0.25, 0.5, 0.5}
\definecolor{codekw}{rgb}{0.35, 0.35, 0.75}
\lstdefinestyle{Pytorch}{
    language = Python,
    backgroundcolor = \color{white},
    basicstyle = \fontsize{9pt}{8pt}\selectfont\ttfamily\bfseries,
    columns = fullflexible,
    aboveskip=1pt,
    belowskip=1pt,
    breaklines = true,
    captionpos = b,
    commentstyle = \color{codeblue},
    keywordstyle = \color{codekw},
}

\definecolor{green}{HTML}{009000}
\definecolor{red}{HTML}{ea4335}

\usepackage{enumitem}
\usepackage[noend]{algpseudocode}
\usetikzlibrary{arrows.meta,calc}
\usepackage{lato}
\definecolor{lstgray}{HTML}{6B7280}
\definecolor{lstblue}{HTML}{3B7FC4}
\definecolor{lstgreen}{HTML}{2E8B63}
\definecolor{lstorange}{HTML}{C9781A}
\lstdefinestyle{suppcode}{basicstyle=\ttfamily\fontsize{7.5}{8.8}\selectfont, columns=fixed, basewidth=0.5em, keepspaces=true,
    showstringspaces=false, breaklines=false, upquote=true, aboveskip=0pt, belowskip=0pt,
    literate={·}{{$\cdot$}}1 {⋅}{{$\cdot$}}1}
\newcommand{\suppbox}[4][]{\tcbinputlisting{enhanced, breakable, listing only, listing file={#4}, listing options={style=suppcode},
    colback=#2!4!white, colframe=#2!75!black, coltitle=white, colbacktitle=#2!75!black, fonttitle=\bfseries\scriptsize, title={#3},
    boxrule=0.5pt, arc=1.5pt, left=3pt, right=3pt, top=2pt, bottom=2pt, toptitle=1.2pt, bottomtitle=1.2pt,
    before skip=5pt, after skip=7pt, #1}}
\providecommand{\Description}[1]{}
\providecommand{\secref}[1]{Section~\ref{sec:#1}}
\providecommand{\figref}[1]{Figure~\ref{fig:#1}}

\title{AgentAssemble: Self-Improving Agents for Articulated 3D Asset Assembly}
\author[1,2]{Lian FU}
\author[1]{Fengbi Lan}
\author[1,2]{Ruihan Yu}
\author[1]{Kaipeng Zhang}
\author[1]{Zhixiang Wang}

\affiliation[1]{Alaya Lab}
\affiliation[2]{The University of Tokyo}

\abstract{
Articulated 3D assets, doors, drawers, faucets, appliances, are essential ingredients for physical simulation, embodied AI training, and interactive content, yet authoring them remains expensive, because every part must be placed at the right pose and scale and every joint must be given the right pivot and axis.
Feed-forward models are trained on a fixed set of categories and struggle with out-of-distribution objects, and procedural scripts must be written by hand for every new category.
LLM agents that operate 3D tools generalize further, but they follow a procedure that a human writes once, so the same mistakes recur on every object.
We instead let this procedure, the agent's \emph{Skill}, improve from its own experience. An LLM-driven agent is given separate part meshes, a reference image, and a kinematic template, and must produce a valid description file.
We present AgentAssemble, a self-improvement framework with two nested loops.
In the inner Runtime Assembly Loop, an Executor agent assembles the parts under the current Skill, while a Runtime Validator script checks intermediate results without ground truth.
In the outer Self-Improvement Loop, a GT-based Evaluator script scores each predicted URDF kinematically, invariant to naming, axis sign, and offset placement, and a Skill Evolver agent rewrites the Skill, including its checks and helper tools, from the resulting diagnostics.
Both loops run on a lightweight, agent-native simulation and rendering engine on which many agents assemble in parallel.
We also introduce \emph{ArtiAssembly}, a benchmark of $300$ articulated objects with ground-truth URDFs and sanitized part meshes.
Experiments show that, through self-improvement alone and without any human edit, AgentAssemble raises the fitness on objects of held-out categories from $0.474$ to $0.681$, and that the improved Skill transfers to other Executor models.
}

\project{\url{https://alaya-lab.github.io/AgentAssemble/}}
\code{\url{https://github.com/AlayaLab/AgentAssemble}}
\correspondence{Zhixiang Wang (Project Lead) and Kaipeng Zhang}
\date{\today}

\begin{document}
\maketitle

\begin{figure}[!h]
    \centering
    \includegraphics[width=\linewidth]{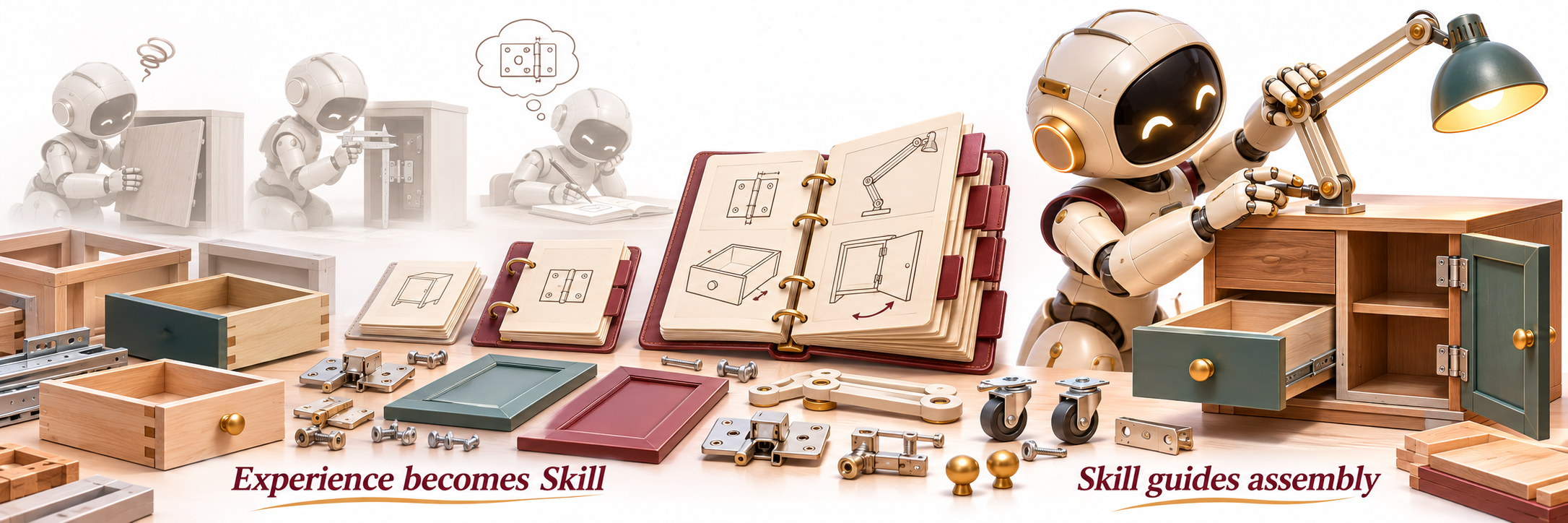}\vspace{5pt}
    \captionof{figure}{\textbf{Experience becomes Skill; Skill guides assembly.}
    \textsc{AgentAssemble} assembles articulated 3D assets from separate parts and refines its assembly procedure and geometric helpers using feedback across repeated rollouts.}
    \label{fig:teaser}
\end{figure}

\section{Introduction}
\label{sec:intro}

Articulated 3D assets, objects whose parts move relative to each other along well-defined kinematic joints, underpin a growing set of applications, from robotic simulation and embodied AI training to interactive 3D content.
For many objects the part meshes already exist, as they do for CAD models, artist-made assets, and the output of part-level generators, and what is missing is their assembly into a working mechanism.
We study this \emph{part-to-object assembly}. Given the part meshes, a reference image, and a template that states the kinematic tree with its joint types and limits, the task is to recover the pose and scale of every part and the pivot and axis of every joint.
A single misplaced pivot silently breaks downstream simulation, so the assembly is still done by hand and limits the supply of simulation-ready assets.

Prior work produces articulated assets along three routes.
Feed-forward models~\cite{caoPhysXAnythingSimulationReadyPhysical2025,liParticulateFeedForward3D2026} learn articulation from a set of training categories, and procedural pipelines~\cite{joshiProceduralGenerationArticulated2025} encode it in per-category scripts.
More recently, agentic systems~\cite{leArticulateAnythingAutomaticModeling2025,shuiArtiCADArticulatedCAD2026,zhangUSDCraftGeometricallyGrounded2026} let an LLM operate a 3D tool, write the articulation as code, and revise it from rendered feedback.
In parallel, recursive self-improvement (RSI) of LLM agents~\cite{zhangDarwinGodelMachine2025,liSkillsBenchBenchmarkingHow2026a} lets an agent improve the harness or skill that produces its own behavior.

None of these routes solves the assembly task well.
Feed-forward models degrade on objects outside their training categories and regenerate the geometry, so the given parts are not preserved, and procedural pipelines need new human authoring for every new category.
Agentic systems generalize further, but the procedure they follow is written once by a human and stays fixed, so the same mistakes recur on every object until a human rewrites it.
Self-improvement could remove this bottleneck, yet three obstacles prevent its direct port to articulated assembly.
First, end-of-rollout judgments are too sparse to tell \emph{which step} of a long assembly procedure went wrong.
Second, an edit that an LLM proposes from feedback is not guaranteed to help, and a harmful edit that is kept degrades the rollouts and the feedback of every later round.
Third, existing RSI and agent-skill benchmarks~\cite{EvoAgentBenchMultiDomainAgent,mengRSIBenchDataBenchmarkingDataCentric2026} do not include 3D assembly, so the data, the engine, and the evaluation that such a loop needs are missing for articulated assets.

We present AgentAssemble, a self-improvement framework in which the assembly procedure, the \emph{Skill}, is rewritten by an agent from feedback on its own rollouts, so that it needs no model training and no per-category code.
An \textbf{Executor} agent follows a Skill and drives a lightweight simulation and rendering engine to assemble the parts into a URDF, with feedback from a \textbf{Runtime Validator} script that checks intermediate assemblies without ground truth.
A \textbf{GT-based Evaluator} script then compares each result kinematically with the ground truth and returns a pass decision, a graded fitness score, and per-joint diagnostics that name the failing check of every joint.
A \textbf{Skill Evolver} agent rewrites the Skill, checks included, from the diagnostics of a batch of rollouts.
Every edit passes a \textbf{two-layer regression strategy}, in which the Skill Evolver first tests the edit on a few train objects and a validation gate then admits the new Skill only if its fitness improves on a separate split.
To supervise and evaluate the framework, we introduce the \emph{ArtiAssembly} benchmark, a set of $300$ procedurally generated articulated objects with ground-truth URDFs.

\paragraph{Contributions.}
In summary, this paper makes three contributions.
\begin{itemize}\setlength\itemsep{2pt}
    \item A \textbf{self-improving agentic assembly framework}, in which an Executor turns separate part meshes and a reference image into a valid URDF, and the Skill it follows, including its checks and helper tools, is rewritten from feedback on its own rollouts.
    \item A \textbf{GT-based Evaluator} and a \textbf{two-layer regression strategy} that make the self-improvement reliable. The Evaluator returns a graded fitness score and per-joint diagnostics, and the regression strategy, a trial on train objects followed by a validation gate, keeps only the edits that improve the Skill.
    \item The \textbf{ArtiAssembly} benchmark with an agent-native simulation and rendering engine, on which the improved Skill generalizes to held-out object categories and transfers to other Executor models.
\end{itemize}

\begin{figure*}[t]
    \centering
    \resizebox{\textwidth}{!}{\lato\input{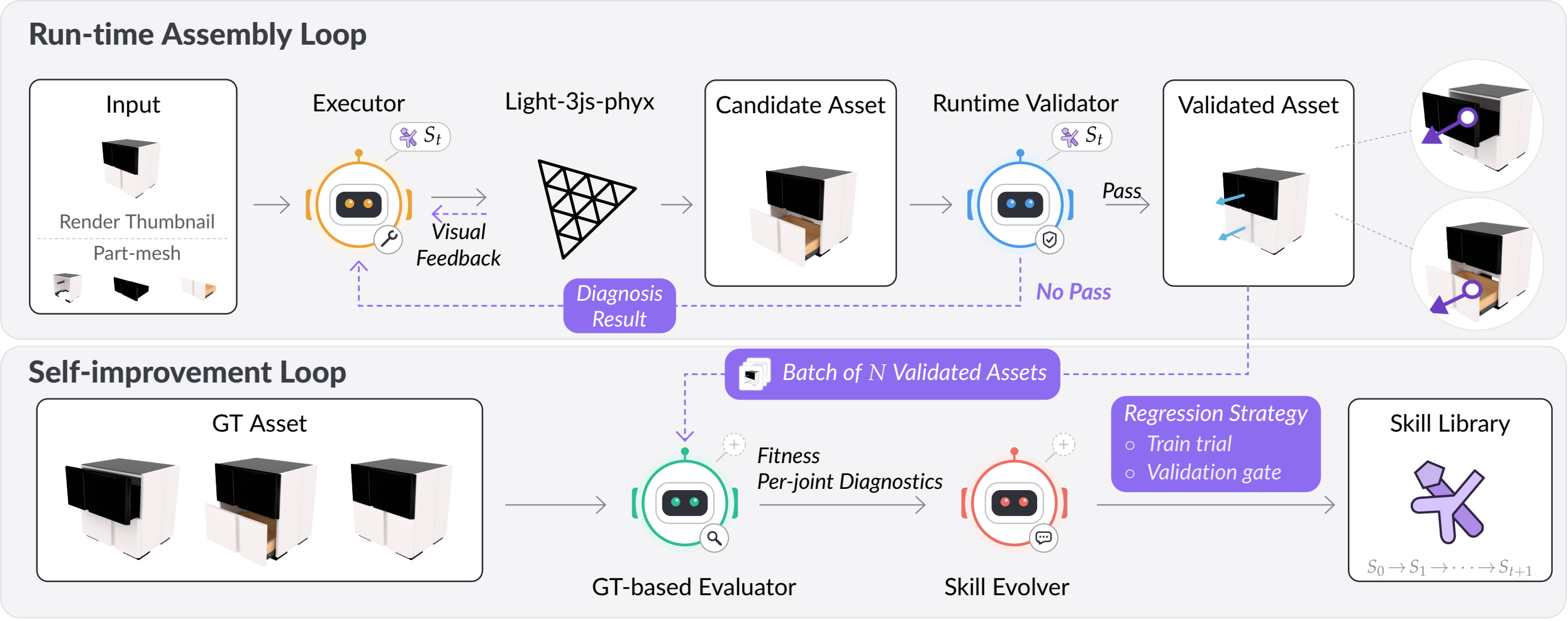}}%
    \caption{Overview of AgentAssemble.
    \textit{Runtime Assembly Loop} (top): the \textbf{Executor} agent reads the current Skill and assembles the parts in the simulation and rendering engine, and the \textbf{Runtime Validator} script checks its intermediate results without ground truth until the URDF is accepted.
    \textit{Self-Improvement Loop} (bottom): the \textbf{GT-based Evaluator} script compares each predicted URDF with the ground truth and outputs the pass decision, the fitness, and the per-joint diagnostics, from which the \textbf{Skill Evolver} agent rewrites the Skill $S_t$.
    The rewritten Skill enters the Skill Library as $S_{t+1}$ only if it passes the two-layer \textbf{regression strategy}.}
    \label{fig:evo_pipeline}
\end{figure*}

\section{Related Work}
\label{sec:related}

\noindent\textbf{Articulated 3D asset generation.}
Most methods learn a mapping from text or images to articulation parameters.
Some produce a simulation-ready URDF in a single forward pass or from a learned articulation prior~\cite{caoPhysXAnythingSimulationReadyPhysical2025,liParticulateFeedForward3D2026,leiNAPNeural3D2023,suArtFormerControllableGeneration2025,chenArtiLatentRealisticArticulated2025a}.
Others condition on single images, dual-state images, or VLM-tagged parts~\cite{liuSINGAPOSingleImage2024,liuCAGEControllableArticulation2024,wuDIPODualStateImages2025,leArticulateAnythingAutomaticModeling2025}, or reconstruct a URDF from one observation~\cite{chenURDFormerPipelineConstructing2024,mandiReal2CodeReconstructArticulated2024,liuPActPartDecomposedSingleView2026}.
Their training data come from a closed taxonomy~\cite{moPartNetLargescaleBenchmark2018,xiangSAPIENSimulAtedPartBased2020a,gengGAPartNetCrossCategoryDomainGeneralizable2023a,liuAKB48RealWorldArticulated2022}, so they degrade on categories that it under-represents.
Procedural generators~\cite{joshiProceduralGenerationArticulated2025,lianInfiniteMobilityScalable2025} and training-free methods~\cite{chenFreeArt3DTrainingFreeArticulated2025a} avoid this at the cost of per-category engineering.
Closer to our input, 3D part assembly methods~\cite{liLearning3DPart2020,liCategoryLevelMultiPartMultiJoint2024} learn to predict the poses of given parts, and they are trained per category and leave the joints of the assembled object unspecified.
In our setting, the parts are given as separate meshes at unknown orientation and scale, the pose of every part and the pivot and axis of every joint must be recovered, and no per-category data or code is available.

\noindent\textbf{Agentic 3D generation.}
Recent work treats 3D generation as an LLM agent that operates a scene editor.
Some systems compose pre-built assets into rooms or layouts~\cite{xiaSAGEScalableAgentic2026,pfaffSceneSmithAgenticGeneration2026a,lingScenethesisLanguageVision2025}, and others reconstruct interactive scenes from video~\cite{xiaHoloSceneSimulationReadyInteractive2025} or generate the interactive parts of a single object~\cite{shanInteract3DCompositional3D2026}.
Closest to ours are ArtiCAD~\cite{shuiArtiCADArticulatedCAD2026}, a multi-agent CAD assembly system with a fixed four-stage loop, and USDCraft~\cite{zhangUSDCraftGeometricallyGrounded2026}, in which an LLM writes and revises a program that compiles a single mesh into an articulated USD asset.
In all of these systems, the procedure that the agent follows is written once by a human and stays fixed.
In ours, the assembly procedure improves itself from the outcomes of its own rollouts.

\noindent\textbf{Recursive self-improvement of LLM agents.}
In recursive self-improvement (RSI), a system uses its current capability to improve the machinery that produces that capability.
Recent LLMs and VLMs have made it practical, since an agent can now read the outcome of its own rollouts and edit the programs that scaffold it~\cite{zelikmanSelfTaughtOptimizerSTOP2023,yinGodelAgentSelfReferential2024,huAutomatedDesignAgentic2024}.
Some frameworks maintain an archive of self-modified coding agents selected by benchmark performance~\cite{zhangDarwinGodelMachine2025,wangHuxleyGodelMachineHumanLevel2025,liuMendelGodelMachine2026,wengGroupEvolvingAgentsOpenEnded2026,zhangHyperagents2026},
and others use LLMs as mutation operators in an evolutionary search over code~\cite{novikovAlphaEvolveCodingAgent2025,langeShinkaEvolveOpenEndedSampleEfficient2025}.
A further group keeps the backbone frozen and improves the \emph{harness} around it~\cite{leeRecursiveHarnessSelfImprovement2026,zhangSelfHarnessHarnessesThat2026,xiaRRSIRegularizedRecursive2026,wuModularRSIModularGeneralizable2026}
or the reusable \emph{skills} that an agent loads at inference time~\cite{alzubiEvoSkillAutomatedSkill2026,zhangCoEvoSkillsSelfEvolvingAgent2026,yangSkillOptExecutiveStrategy2026,liuSkillVineAgentSkill2026,wangMetaSkillEvolveRecursiveSelfImprovement2026,liDynamicAgentSkills2026}.
These frameworks have been studied mostly on coding, reasoning, and text-based tasks.
First attempts move RSI into spatial reasoning and embodied settings~\cite{liSpatialEvoSelfEvolvingSpatial2026,juEmbodiSkillSkillAwareReflection2026,wangRegenHarnessRobotAgent2026,qinEmbodiedSmithScalingEmbodied2026}, and none of them addresses the assembly of 3D assets.
We explore the skill-level form of this line for articulated 3D assets, where the agent works from meshes and images and its output is judged by kinematics.
\section{Method}
\label{sec:method}

\subsection{Task Formulation}
\label{sec:method:problem}

\paragraph{Input and output.}
An object instance is a triple $x = (\{M_i\}_{i=1}^{n}, I, U^{0})$ of $n$ part meshes $M_i$, a single reference image $I$ of the assembled object at rest, and a URDF template $U^{0}$.
The task is to produce an articulated asset encoded as a URDF, $U = (\mathcal{L}, \mathcal{J})$.
The links $\mathcal{L}$ are in one-to-one correspondence with the parts, and the joints $\mathcal{J}$ form a rooted tree over them, each joint $j$ being a tuple
\[
j = \big(\mathrm{parent}_j,\ \mathrm{child}_j,\ \mathrm{type}_j,\ \mathbf{p}_j,\ \mathbf{a}_j,\ [q^{\mathrm{lo}}_j, q^{\mathrm{hi}}_j]\big),
\]
where $\mathrm{type}_j$ is one of \emph{revolute}, \emph{continuous}, \emph{prismatic}, and \emph{fixed}, $\mathbf{p}_j \in \mathbb{R}^3$ is the pivot, $\mathbf{a}_j \in S^2$ is the axis direction, and the motion limits contain the rest configuration $q_j = 0$.
The benchmark poses the task as \emph{joint filling}.
The template $U^{0}$ gives the structural fields of $U$, namely the link tree with $\mathrm{parent}_j$ and $\mathrm{child}_j$, the link-to-part mapping, the joint types, the motion limits, and any mimic couplings between joints.
It leaves the geometric fields blank, and these are what the agent must recover.
For every non-root link $i$ they are the rotation $R_i \in SO(3)$ and the uniform scale $s_i > 0$ that return its mesh to the assembled orientation and metric size, and for every joint they are the pivot $\mathbf{p}_j$ in the parent frame and the world-space axis $\mathbf{a}_j$.
The URDF defines forward kinematics $T_i(\mathbf{q}) \in SE(3)$ for every link as a function of the joint vector $\mathbf{q}$.
At $\mathbf{q} = \mathbf{0}$ it must reproduce the rest pose shown in $I$, and driving each joint through its given limits must move the parts as the real object does.
Recovering these quantities from the part geometry and the reference image alone is what we call \emph{assembling} the object.

\paragraph{Correctness.}
A ground-truth URDF $U^\ast_x$ exists for every object, but $U$ and $U^\ast_x$ cannot be compared as text, because the same mechanism admits many encodings that differ in names, in the sign of an axis together with its limits, or in where an offset is stored.
We therefore compare the two URDFs by the motion they produce, after matching every link of $U$ to its ground-truth counterpart by geometry.
We call $U$ \emph{correct} when it meets three conditions.
First, its structure equals that of $U^\ast_x$, i.e.\ the same link tree and the same joint types and limits.
Second, at the rest configuration $\mathbf{q} = \mathbf{0}$ every link lies within a tolerance $\tau$ of its ground-truth counterpart, which requires the right orientation, size, and placement of the part.
Third, when each joint is driven through its limits, the moving links stay within $\tau$ of their ground-truth trajectories, which requires the right pivot and axis.
The comparison yields three outputs.
\begin{itemize}
    \item \emph{Pass}: $\Phi(U, U^\ast_x) \in \{0, 1\}$ is one when $U$ is correct, i.e.\ when every check passes, and zero otherwise, so an assembly with a single misplaced joint counts as a failure.
    The \emph{Pass Rate} is the fraction of rollouts with $\Phi = 1$, i.e.\ the share of assemblies that can be used as they are.
    \item \emph{Fitness Score}: $\phi(U, U^\ast_x) \in [0, 1]$ gives partial credit per joint.
    A joint earns credit for being correct and further credit for how close its trajectory is to the ground truth, so $\phi$ rises as an assembly improves even while $\Phi$ stays at zero.
    \item \emph{Per-joint Diagnostics}: $\mathcal{E}(U, U^\ast_x)$ name the offending dimension of each joint.
    For every joint they record whether its axis, pivot, and trajectory are within tolerance and how large each error is, and they are the feedback from which the Skill is improved.
\end{itemize}

\paragraph{Agentic assembly.}
We solve the task with an agent and without training a model. A fixed multi-modal LLM acts as an agent that works through tool calls to a simulator and finally exports a URDF.
The agent's behavior is conditioned on a \emph{Skill} $S$, so the agent together with the Skill defines a stochastic policy $\pi_S$ with $U \sim \pi_S(x)$.

\paragraph{Skill.}
A Skill $S$ is a directory.
It holds a Markdown procedure that tells the agent how to assemble an object, deterministic helper scripts with their documentation and a preflight test, and the contract of the runtime the agent calls.
The whole Skill is loaded into the agent's context at the start of every rollout, and it is the only component of the system that changes between rounds, whereas the backbone LLM, the simulation runtime, and the evaluation are frozen.
We write $S_t$ for the candidate Skill that is proposed in round $t$ of self-improvement and call the hand-written Skill that the loop starts from the seed Skill $S_0$.

\paragraph{Skill optimization.}
We optimize the Skill and keep the weights of the LLM fixed,
\[
S^\ast = \arg\max_S\ \mathbb{E}_{x \sim \mathcal{D}}\ \mathbb{E}_{U \sim \pi_S(x)}\ \big[\Phi(U, U^\ast_x)\big],
\]
where $\mathcal{D}$ is the distribution of objects.
Since $\Phi$ is all-or-nothing and too coarse to compare two Skills on a small set of objects, the search uses the graded score $\phi$ as its surrogate.
The search is gradient-free and is carried out by a second LLM that edits $S$ from the diagnostics $\mathcal{E}$ on a small training set of objects with ground truth.
The resulting Skill is then evaluated on held-out test objects.

\subsection{Framework Overview}
\label{sec:method:framework}
AgentAssemble optimizes the Skill objective through a four-role pipeline, as shown in \figref{evo_pipeline}.
The Executor and Skill Evolver are LLM agents sharing a fixed multi-modal backbone. The Runtime Validator and GT-based Evaluator are deterministic scripts.
The \emph{Runtime Assembly Loop} runs once per object, in which the Executor alternates assembly actions with calls to the Runtime Validator, revising the candidate from its checking results until the checks pass and the final URDF is accepted.
The \emph{Self-Improvement Loop} runs once per round over a batch of training objects, in which every object is rolled out through the inner loop, the Evaluator scores each session against the ground truth, and the Skill Evolver proposes a revised Skill for the next round.

\subsection{Runtime Assembly Loop}
\label{sec:method:skill}

\paragraph{Executor.}
The Executor is the LLM agent implementing the policy $\pi_S$.
It reads the Skill $S$ and one object $x$, and acts through tool calls to a persistent Python session, into which the SDK of our simulation and rendering engine and a frozen \emph{Assembly SDK} are preloaded.
In this session it imports and inspects the parts, authors links and joints, drives joints through their range, renders the scene, and finally exports the URDF $U$ together with the rollout artifacts.
It invokes the Runtime Validator at the checkpoints prescribed by the Skill and uses the returned checking results to refine the assembly.

\paragraph{Runtime Validator.}
The Runtime Validator is a script-based role that checks intermediate and candidate assemblies without access to ground truth.
It combines the runtime's structural audit with the deterministic helpers the Skill ships.
It returns checking results to the Executor, which decides (1) whether the assembly needs to be revised and (2) how to revise it.
The checking results turn the sparse end-of-rollout signal into detailed feedback inside the rollout.
The Runtime Validator is part of the Skill, called by the Executor, and can be improved by the Skill Evolver.

\subsection{Self-Improvement Loop}
\label{sec:method:skill_loop}

\begin{figure*}[t]
    \centering
    \includegraphics[width=\textwidth]{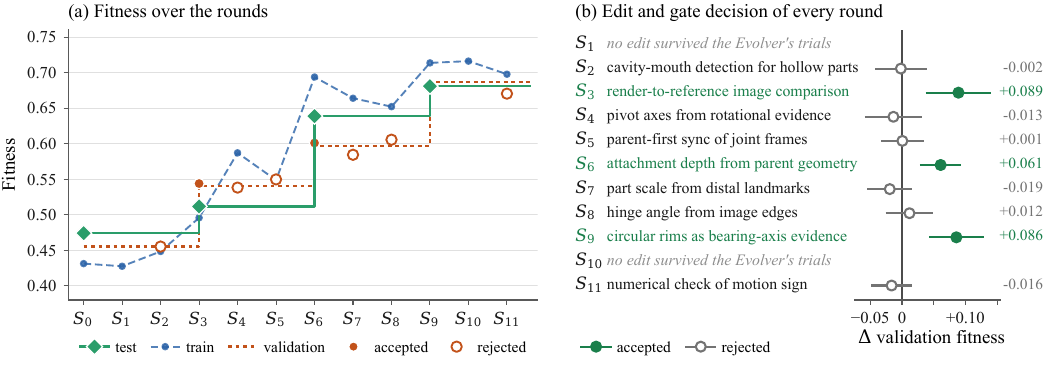}
    \caption{Self-improvement of the Skill from the seed Skill $S_0$ to the last candidate $S_{11}$.
    $S_t$ is the candidate Skill of round $t$.
    (a)~Fitness on the three splits.
    Test: the champion after each round on the test split, with a diamond at the seed Skill $S_0$ and at each accepted Skill.
    Train: each candidate on the train split.
    Validation: the champion after each round, pooled over its validation sessions, with a dot for each candidate.
    (b)~The edit that the Skill Evolver proposed in every round and the decision of the validation gate.
    The dot is the mean per-object difference in validation fitness between the candidate and the champion, and the whisker spans $1.5$ standard errors on each side.
    A candidate is accepted when its whisker stays to the right of zero.}
    \label{fig:evolution}
\end{figure*}

\paragraph{GT-based Evaluator.}
After export, the Evaluator compares the predicted URDF $U$ with the ground truth $U^\ast_x$ using three checks.
The \emph{structural check} $c^{\mathrm{str}} \in \{0,1\}$ tests that $U$ is a valid tree whose links all match the ground truth.
Let $\mathcal{J}^\ast$ denote the set of movable joints of the ground truth $U^\ast_x$. For every joint $j \in \mathcal{J}^\ast$, the \emph{articulation check} $c^{\mathrm{art}}_j \in \{0,1\}$ tests the type, the axis direction, and the limits of the joint.
The \emph{geometric check} $c^{\mathrm{geo}}_j \in \{0,1\}$ tests that the rest pose, the pivot, and the trajectory of the moving links over a sweep of the joint lie within a tolerance $\tau = \max(2\,\text{cm}, 0.04\,d)$ of the ground truth, in which $d$ is the diagonal of the ground-truth object.
A joint is correct when all three checks pass,
\begin{equation}
    c_j = c^{\mathrm{str}} \, c^{\mathrm{art}}_j \, c^{\mathrm{geo}}_j .
    \label{eq:joint_flag}
\end{equation}
The Evaluator also records the drift $\delta_j$ of the trajectory from the ground truth and the measured errors $\mathbf{e}_j$ behind the checks, such as the axis angle and the pivot distance.
These records form the diagnostics
\begin{equation}
    \mathcal{E}(U, U^\ast_x) = \Big( c^{\mathrm{str}},\ \big\{ (c^{\mathrm{art}}_j,\ c^{\mathrm{geo}}_j,\ \delta_j,\ \mathbf{e}_j) \big\}_{j \in \mathcal{J}^\ast} \Big).
    \label{eq:diagnostics}
\end{equation}
The graded score gives each joint half of its credit for being correct and half for being close,
\begin{equation}
    \phi(U, U^\ast_x) = \frac{1}{|\mathcal{J}^\ast|} \sum_{j \in \mathcal{J}^\ast} \Big[ \tfrac{1}{2}\, c_j + \tfrac{1}{2}\, \mathrm{clip}\big(1 - \tfrac{\delta_j}{3\tau},\, 0,\, 1\big) \Big],
    \label{eq:fitness}
\end{equation}
and $\phi = 0$ when no URDF is exported or a structural check fails.
We build the Evaluator from deterministic checks, whereas other work uses agents as judges~\cite{linProceduraAgentic3D2026}, because a model that grades its own output can be biased.
The Evaluator is the only place where ground truth enters the system, and its outputs are redacted of ground-truth values before any agent sees them.

\paragraph{Skill Evolver.}
The Skill Evolver is a code-writing agent that turns the Skill $S_t$ into $S_{t+1}$.
It receives the Evaluator's diagnostics $\{\mathcal{E}(U_k, U^\ast_{x_k})\}_{k \in \mathcal{B}_t}$ of the batch of rollouts $\mathcal{B}_t$ of round $t$, together with the rollout artifacts, reads the current Skill, and rewrites it.
It may edit every component of the Skill, and everything else stays frozen, which keeps rollouts comparable across rounds.
An audit of its transcript after every round confirms that it reads no ground-truth file.

\paragraph{Self-improvement with Regression Strategy.}
The self-improvement process keeps a \emph{champion} $S^\star$, the best Skill found so far, which starts as the seed Skill $S_0$.
In each round, the diagnostics of the train rollouts are grouped into clusters of failures that share a joint type and an error pattern, and the Skill Evolver works on the $N_c$ largest clusters, which hold the most recurring errors.

The Skill Evolver forks a candidate $S_t$ from the champion $S^\star$ and edits it based on the train rollouts of the previous candidate, which differs from the champion by the edits of at most one round, so its failures remain informative.
In practice, $S_t$ is not guaranteed to improve over $S^\star$, and a worse Skill, once deployed, may break the rollouts and the feedback of the next round, which would make the loop diverge.
We therefore adopt a two-layer regression strategy.
First, the Skill Evolver tests its own edits on $N_o$ train objects of the failure cluster before accepting them, so that it only proposes edits that improve the fitness on the train objects.
Otherwise it takes the feedback and tries again, with at most $M$ attempts per cluster.
Each edit is compared with the Skill before that edit on the same $N_o$ objects.
An edit that fails all $M$ attempts is discarded, and the edits that are kept for the different clusters accumulate in the same candidate $S_t$.
A candidate $S_t$ that is better on the train objects may still overfit to them through case-specific edits.
The second layer of regression is the \emph{validation gate}.
$S_t$ is run on a separate validation split with $K$ sessions per object, and the candidate is accepted if the mean per-object difference in fitness between the candidate and the champion exceeds zero by a fixed number of standard errors.
A rejected candidate is discarded and the next round forks from the same champion.
The separate split guards against overfitting to the train objects, and the $K$ sessions guard against accepting an edit on the noise of a single rollout.
Only if a candidate passes both layers of regression is it accepted as the new champion $S^\star$ for the next round.
The two-layer regression strategy thus keeps harmful edits out of the champion at a low cost.

\section{Experiments}
\label{sec:results}

\subsection{Experimental Setup}
\label{sec:exp:setup}

\noindent\textbf{Dataset and splits.}
We introduce the \emph{ArtiAssembly} benchmark, a set of $300$ articulated everyday objects in four tiers of planned complexity, generated with the agent-driven method Procedura~\cite{linProceduraAgentic3D2026}.
Every object belongs to a different object category, e.g.\ laptop, cash register, or dishwasher, across $26$ domains.
Table~\ref{tab:datasets} compares the dataset with existing ones.
For each object, we merge the geometry into one mesh per link, replace every semantic name and material with a neutral index and a flat color, rotate and rescale each non-root mesh at random, and blank the joint origins, joint axes, link orientations, and link scales in the URDF template that the agent must fill.
To bound the cost of the rollouts, we run the study on a $120$-object subset selected with the seed Skill $S_0$ before any self-improvement.
We run $S_0$ with $K = 3$ sessions on the $213$ objects that have at most eight joints and whose links are all visible in the reference image, where it reaches a fitness of $0.538$ and a Pass Rate of $37.2\%$.
The subset is drawn from the $105$ objects that $S_0$ never solves, which leave room for improvement, and the $53$ that it always solves, which reveal regressions.
It is split at random into train, validation, and test sets of $40$ objects each, used for the feedback to the Skill Evolver, the validation gate, and the final evaluation, and every split holds $32$ objects of the first group and $8$ of the second.
All results that we report come from new sessions, so the screening sessions are never reused.

\begin{figure*}[t]
    \centering
    \includegraphics[width=0.98\textwidth]{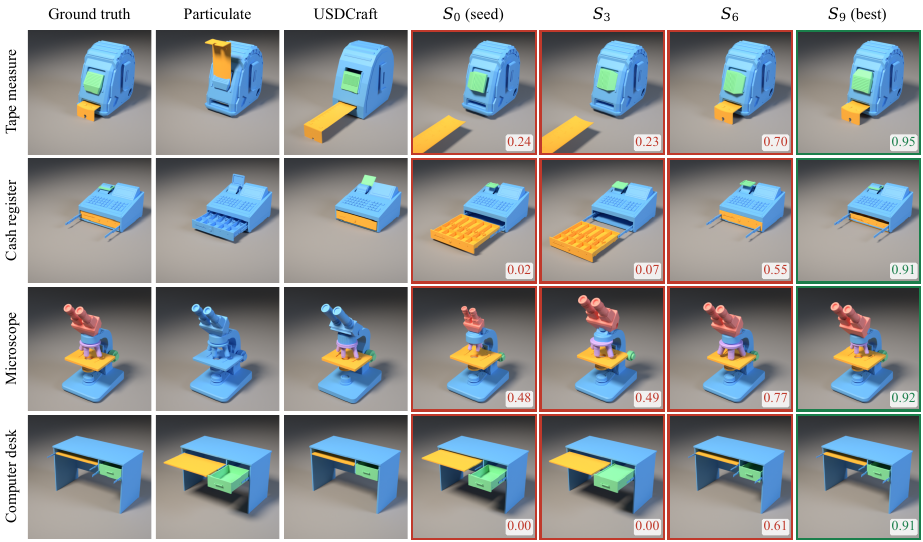}
    \caption{Qualitative results of the two existing methods, the seed Skill $S_0$, and the accepted Skills.
    Particulate and USDCraft take the assembled mesh as input, and each of their parts is drawn in the color of the ground-truth link that it covers.
    The number in each cell is the fitness of the session. During self-improvement, the Skills gradually get better at assembling the parts.}
    \label{fig:qualitative}
\end{figure*}

\begin{table}[htb]
    \centering
    \caption{ArtiAssembly and existing sets of articulated objects.
    {Category: number of object categories, where every object of ArtiAssembly belongs to its own category.
    Tiers: complexity tiers defined.
    Code: parametric source code of the geometry released.}}
    \label{tab:datasets}
    \tablestyle{3pt}{1.1}
    \begin{tabular}{@{}lrrccl@{}}
    \shline
    Dataset & Obj. & Category & Tiers & Code & Format \\
    \hline
    PartNet-Mobility~\cite{xiangSAPIENSimulAtedPartBased2020a} & 2346 & 46 & $\times$ & $\times$ & URDF \\
    ArtVIP~\cite{jinArtVIPArticulatedDigital2026} & 992 & 37 & $\times$ & $\times$ & USD \\
    Lightwheel~\cite{LightwheelAILightwheelsimreadyassetOpensource} & 259 & 14 & $\times$ & $\times$ & USD \\
    \hline
    \textbf{ArtiAssembly} & \textbf{300} & \textbf{300} & \checkmark & \checkmark & \textbf{URDF} \\
    \shline
    \end{tabular}
\end{table}

\noindent\textbf{Implementation.}
The two LLM agents, the Executor and the Skill Evolver, run on \texttt{gpt-6.1-sol}.
The Executor runs at low reasoning effort with a budget of $60$ tool turns per rollout, and the Skill Evolver runs at the highest effort.
Every round rolls out the train objects with one Executor session per object under the current candidate Skill, and every candidate is then evaluated on the validation objects with $K = 3$ independent sessions per object.
In every round, the Skill Evolver works on the $N_c = 3$ largest failure clusters, and for each of them it has $M = 2$ attempts to find an edit that improves the fitness on $N_o = 3$ train objects of the cluster.
The validation gate accepts a candidate when its mean per-object fitness exceeds that of the champion by at least $1.5$ standard errors.
The run consists of twelve rounds of train rollouts, the first under the seed Skill $S_0$, so the Skill Evolver proposes eleven candidates, $S_1$ to $S_{11}$, and it receives no human edit or comment.
After the run, we evaluate the seed Skill $S_0$ and every accepted Skill on the test objects, again with $K = 3$ sessions per object.
For simulation and rendering, we implement a lightweight engine named \emph{Light-3js-phyx}, which is based on three.js and Rapier.
The engine produces pixel-aligned RGB, per-link segmentation, and metric depth, and provides a Model Context Protocol (MCP) interface through which the agents control the simulation and rendering.

\subsection{Self-Improvement on ArtiAssembly}
\label{sec:exp:skill}

We run the Self-Improvement Loop with the settings above, starting from the seed Skill $S_0$ that states only the input, the output, and the SDK.
\figref{evolution}a plots the twelve rounds on the three splits, and \figref{evolution}b lists the edit and the gate decision of every round.

\noindent\textbf{Metrics.}
We report \emph{Pass Rate}, the mean of the pass predicate $\Phi$, and \emph{Fitness}, the mean of the graded score $\phi$ of Eq.~\eqref{eq:fitness}, both computed by the GT-based Evaluator.
Both are averaged first over the $K$ Executor sessions of an object and then over objects.

\begin{table*}[t]
    \centering
    \caption{Comparison on the ArtiAssembly test split.
    Easy objects have one movable joint, medium objects two or three, and hard objects four or more, and the number of test objects is given in parentheses.
    Jnt.: correct joints.
    Ax., Piv., Mot.: joints correct in axis, pivot, and motion.
    Obj.: objects with all joints correct.
    All values are percentages.
    Model: the LLM that the method calls, where astra is \texttt{gpt-6-astra} and sol is \texttt{gpt-6.1-sol}.
    $S_{\mathrm{best}}$ is the final champion of the self-improvement, $S_9$.}
    \label{tab:comparison}
    \tablestyle{4pt}{1.1}
    \resizebox{\textwidth}{!}{%
    \begin{tabular}{@{}ll|ccccc|ccccc|ccccc|ccccc@{}}
    \shline
    & & \multicolumn{5}{c|}{Easy (11)} & \multicolumn{5}{c|}{Medium (15)} & \multicolumn{5}{c|}{Hard (14)} & \multicolumn{5}{c}{All (40)} \\
    Method & Model & Jnt.$\uparrow$ & Ax.$\uparrow$ & Piv.$\uparrow$ & Mot.$\uparrow$ & Obj.$\uparrow$ & Jnt.$\uparrow$ & Ax.$\uparrow$ & Piv.$\uparrow$ & Mot.$\uparrow$ & Obj.$\uparrow$ & Jnt.$\uparrow$ & Ax.$\uparrow$ & Piv.$\uparrow$ & Mot.$\uparrow$ & Obj.$\uparrow$ & Jnt.$\uparrow$ & Ax.$\uparrow$ & Piv.$\uparrow$ & Mot.$\uparrow$ & Obj.$\uparrow$ \\
    \hline
    Particulate~\cite{liParticulateFeedForward3D2026} & -- & 63.6 & 72.7 & 75.0 & 63.6 & 63.6 & 19.4 & 30.6 & 23.1 & 19.4 & 13.3 & 12.8 & 24.4 & 8.8 & 14.1 & 0.0 & 19.2 & 30.4 & 18.7 & 20.0 & 22.5 \\
    USDCraft~\cite{zhangUSDCraftGeometricallyGrounded2026} & astra & \textbf{100.0} & \textbf{100.0} & \textbf{100.0} & \textbf{100.0} & \textbf{100.0} & 91.7 & 91.7 & 96.2 & 91.7 & 80.0 & 76.9 & 78.2 & 71.9 & 76.9 & 50.0 & 83.2 & 84.0 & 81.3 & 83.2 & 75.0 \\
    \hline
    Ours, $S_0$ & sol & 97.0 & 97.0 & \textbf{100.0} & 97.0 & 97.0 & 80.6 & 93.5 & 87.2 & 83.3 & 62.2 & 70.9 & 89.7 & 69.6 & 73.1 & 35.7 & 76.0 & 91.5 & 77.3 & 78.1 & 62.5 \\
    Ours, $S_{\mathrm{best}}$ & sol & \textbf{100.0} & \textbf{100.0} & \textbf{100.0} & \textbf{100.0} & \textbf{100.0} & 91.7 & 93.5 & 92.3 & 92.6 & 86.7 & 86.3 & 93.2 & 83.6 & 87.2 & 73.8 & 89.1 & 93.9 & 87.5 & 89.9 & 85.8 \\
    \hline
    Ours, $S_0$ & astra & \textbf{100.0} & \textbf{100.0} & \textbf{100.0} & \textbf{100.0} & \textbf{100.0} & 85.2 & 96.3 & 89.7 & 85.2 & 71.1 & 77.4 & 92.7 & 77.2 & 80.3 & 45.2 & 81.6 & 94.4 & 82.8 & 83.5 & 70.0 \\
    Ours, $S_{\mathrm{best}}$ & astra & \textbf{100.0} & \textbf{100.0} & \textbf{100.0} & \textbf{100.0} & \textbf{100.0} & \textbf{100.0} & \textbf{100.0} & \textbf{100.0} & \textbf{100.0} & \textbf{100.0} & \textbf{89.7} & \textbf{94.4} & \textbf{90.6} & \textbf{90.6} & \textbf{78.6} & \textbf{93.6} & \textbf{96.5} & \textbf{94.1} & \textbf{94.1} & \textbf{92.5} \\
    \shline
    \end{tabular}}
\end{table*}

\noindent\textbf{Train.}
On the train split, the fitness of the candidate rises from $0.431$ under the seed Skill $S_0$ to $0.698$ in the last round.

\noindent\textbf{Validation and selection.}
The gate compares every candidate with the champion on the validation split and accepted three of the nine edited candidates, $S_3$, $S_6$, and $S_9$, each by a paired gain of $0.06$ to $0.09$ fitness, as shown in \figref{evolution}b.
The three accepted edits are generic geometric procedures.
The gate rejected six candidates whose difference from the champion lay between $-0.02$ and $+0.01$, and two rounds produced no edit.
The validation fitness of the champion, pooled over all of its validation sessions, rises in three steps from $0.455$ to $0.687$, and the run returns the last champion, $S_9$.

\noindent\textbf{Test.}
On the test split, also shown in \figref{evolution}a, $S_9$ raises fitness over the seed Skill $S_0$ by $0.207 \pm 0.041$ (mean and standard error over objects) and Pass Rate from $30.0\%$ to $52.5\%$, and its test result agrees with its validation estimate.
The gain of the selected Skill thus holds on object categories that the loop never saw.

\noindent\textbf{Qualitative results.}
\figref{qualitative} shows four test objects.
Under the seed Skill $S_0$, parts are left at the wrong size, orientation, or position, e.g.\ the blade of the tape measure is stretched and detached, and the drawers of the cash register and the computer desk hang far outside their housing.
The accepted Skills remove these errors step by step, and $S_9$ passes on all four.

\subsection{Transfer across Executor Models}
\label{sec:exp:models}

\begin{figure}[t]
    \centering
    \includegraphics[width=0.62\linewidth]{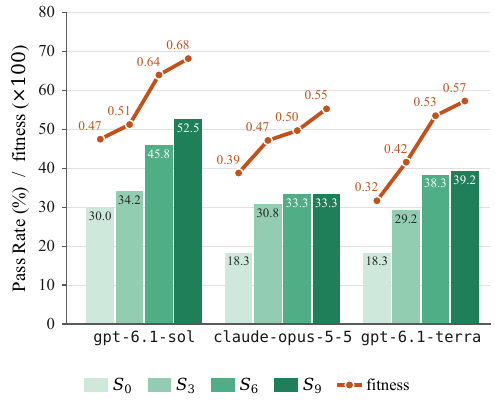}
    \caption{Model transfer.
    The seed Skill $S_0$ and the three accepted Skills are run unchanged on the test split with three Executor models.
    Bars are Pass Rate and the line is fitness.}
    \label{fig:models}
\end{figure}

This experiment tests whether the improved Skill also helps Executor models that took no part in the self-improvement.
We run the seed Skill $S_0$ and the three champions unchanged on the test split with \texttt{claude-opus-5-5} and \texttt{gpt-6.1-terra} as the Executor, at the same reasoning effort and turn budget.
As shown in \figref{models}, fitness rises with every champion under both models.
$S_9$ raises the fitness of \texttt{claude-opus-5-5} by $0.165 \pm 0.034$ and its Pass Rate from $18.3\%$ to $33.3\%$, and it raises the fitness of \texttt{gpt-6.1-terra} by $0.256 \pm 0.039$ and its Pass Rate from $18.3\%$ to $39.2\%$.

\subsection{Comparison with Existing Methods}
\label{sec:exp:comparison}

We compare with two methods that articulate a given 3D object, the feed-forward model Particulate~\cite{liParticulateFeedForward3D2026} and another agentic system USDCraft~\cite{zhangUSDCraftGeometricallyGrounded2026}, which we run with \texttt{gpt-6-astra} at low reasoning effort, as in the original paper.
The two methods take the assembled mesh as input and predict the part segmentation, the kinematic tree, and the joint types themselves, whereas our Executor receives the separated parts at scrambled pose and scale together with the template.
The inputs therefore differ, and the comparison serves as a reference on the same objects.

\noindent\textbf{Metrics.}
Since the two methods rebuild or re-segment the geometry, we match the predicted links to the ground-truth links by geometric overlap at the rest pose, and score every ground-truth movable joint on quantities that are independent of the predicted geometry.
\emph{Axis} requires the joint axis to lie within $10^\circ$ of the ground truth.
\emph{Pivot}, reported for revolute joints, requires the distance between the two axis lines to be at most the tolerance $\tau$ of the GT-based Evaluator.
\emph{Motion} requires the drift of the moving links over the ground-truth joint range to be at most $\tau$.
A joint is correct when its type and all of the above are correct, and an object is correct when all of its joints are.
A missing prediction counts as incorrect, and the joint type is given to our method by the template.
These metrics score the joint parameters only, whereas the pass predicate $\Phi$ also requires every part to lie within $\tau$ of the ground truth at rest and along the motion, which is why \emph{Obj.} exceeds the Pass Rate.

\noindent\textbf{Results.}
As shown in Table~\ref{tab:comparison}, with \texttt{gpt-6-astra}, the model that USDCraft uses, the seed Skill $S_0$ recovers fewer correct objects than USDCraft, and $S_9$ reaches $93.6\%$ of the joints and $92.5 \pm 3.9\%$ of the objects correct, against $75.0 \pm 6.9\%$ of the objects for USDCraft.
The gain grows with difficulty, and on the hard objects $S_9$ raises the share of fully correct objects from $45.2\%$ under $S_0$ to $78.6\%$, against $50.0\%$ for USDCraft.
$S_9$ was improved with \texttt{gpt-6.1-sol} only, with which it also exceeds USDCraft in every column over all objects.

\subsection{Ablation Study}
\label{sec:exp:ablation}

\begin{table}[t]
    \centering
    \caption{Ablation study on the test split.
    Pass Rate is a percentage.
    $\Delta$ is the paired fitness gain over the seed Skill $S_0$, given as mean and standard error over objects.}
    \label{tab:ablation}
    \tablestyle{5pt}{1.1}
    \begin{tabular}{@{}lccc@{}}
    \shline
    Skill & Fitness$\uparrow$ & Pass Rate$\uparrow$ & $\Delta$ \\
    \hline
    Seed Skill $S_0$ & 0.474 & 30.0 & -- \\
    Human-curated Skill & 0.541 & 37.5 & $+0.067 \pm 0.024$ \\
    \hline
    \emph{w/o} regression strategy & 0.475 & 30.0 & $+0.001 \pm 0.032$ \\
    \emph{w/o} diagnostics & 0.609 & 47.5 & $+0.135 \pm 0.030$ \\
    Full method ($S_9$) & \textbf{0.681} & \textbf{52.5} & $\mathbf{+0.207 \pm 0.041}$ \\
    \shline
    \end{tabular}
\end{table}

Table~\ref{tab:ablation} compares the full method with two variants of the self-improvement and with a human-curated Skill, all evaluated on the test split.

\noindent\textbf{Regression strategy.}
We repeat the run with the regression strategy removed, so that every edit of the Skill Evolver is kept and used in the next round.
The Skill of round $9$ falls back to the level of the seed Skill $S_0$, as edits that do not help are kept and accumulate over the rounds.
The regression strategy is therefore what turns unreliable edits into a steady improvement.

\noindent\textbf{Diagnostics of the Evaluator.}
We repeat the run with the Skill Evolver receiving only the fitness of each rollout, without the per-joint diagnostics $\mathcal{E}$ of Eq.~\eqref{eq:diagnostics}.
The regression strategy still filters the edits, so the final champion improves over $S_0$, and it stays below the full method by $0.072 \pm 0.046$ fitness in the paired comparison, a gap that is suggestive at this sample size.
The diagnostics tell the Evolver which joint failed and in which check, so that its edits can address the actual failures.

\noindent\textbf{Human-curated Skill.}
This Skill was written by a human while testing the agent on this task.
It improves over $S_0$ by a small margin and stays below $S_9$ by $0.140 \pm 0.043$ fitness, so the self-improvement finds procedures beyond those of the human-curated Skill.

\section{Conclusions}
\label{sec:conclusions}

We presented AgentAssemble, a self-improvement framework that assembles articulated 3D assets from part meshes, together with the \emph{ArtiAssembly} benchmark.
Starting from a seed Skill without an assembly procedure, the Self-Improvement Loop wrote one with geometric helper tools and no human edit.
The resulting Skill raises the test fitness from $0.474$ to $0.681$ on held-out object categories, and it also improves assembly with Executor models that took no part in the self-improvement.
With the Executor model of USDCraft, it recovers all joints of $92.5\%$ of the test objects, against $75.0\%$ for this baseline.
The GT-based Evaluator and the two-layer regression strategy are what make this improvement reliable, since the diagnostics direct the edits and only the edits that help are kept.
Our method has limitations, as an assembly under $S_9$ takes about two minutes and the result is bounded by the capability of the underlying LLM.
The loop also needs objects with ground-truth URDFs for its train and validation splits.
Future work will extend the framework to harder 3D generation tasks, and will reduce the cost of the self-improvement and of every assembly.

\section*{LLM Usage Statement}
LLMs are a component of the proposed method and the models are named in the experimental setup.
Beyond the method, we used an LLM-based coding assistant (Claude) to edit and polish the text of all sections, to write and debug experiment and plotting code, and to check the metadata of the references.
An image generation model (gpt-image-2) was used to draft the illustration in the teaser figure.
The research idea, the experimental design, and all claims were determined and verified by the authors, who take full responsibility for the content of the paper.

\bibliographystyle{abbrv}
\bibliography{references}

\clearpage
\beginappendix

\section{Task Input}
\label{sec:supp:input}

\begin{figure*}[!htb]
    \centering
    \includegraphics[width=\textwidth]{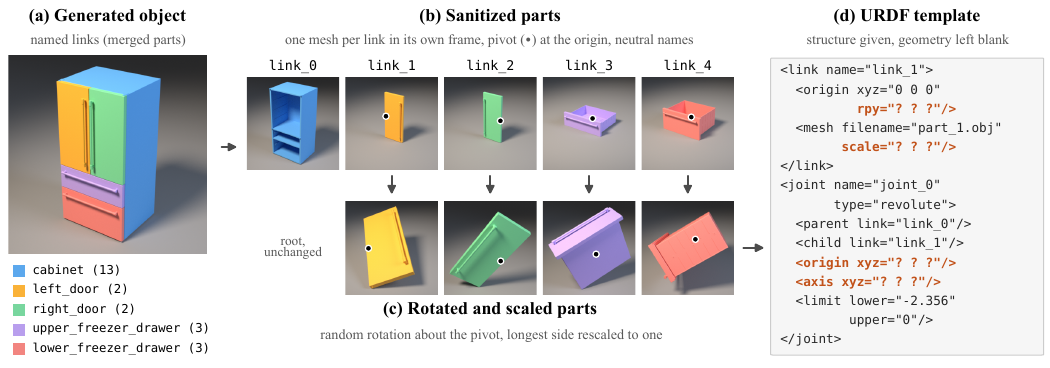}
    \caption{One instance of \emph{ArtiAssembly}, from the generated object to the task that the agent receives.
    (a)~The generated object, a french-door refrigerator with five named links.
    (b)~Sanitization merges the geometry into one mesh per link in the link's own frame and replaces the names with neutral indices.
    (c)~Every non-root part is rotated at random about its pivot and rescaled so that its longest side is one.
    (d)~The URDF template gives the structure and leaves the rotation and scale of every part and the origin and axis of every joint blank.
    The reference image of the task shows the object of panel (a) at rest with one flat color per link.}
    \label{fig:dataset_sample}
\end{figure*}

This section describes what the agent receives for one object, as formulated in \secref{method:problem}.
\figref{dataset_sample} follows one instance from the generated object to the task files.

\paragraph{Files of a task.}
Each instance presents the agent with a directory of part meshes, a URDF template, and a single reference image of the assembled object at rest, rendered from the ground-truth URDF with the same headless renderer the Executor uses.
The ground truth lives in a separate tree that only the GT-based Evaluator may read.
All identifiers are \emph{sanitized} before the agent sees them, so that categories become \texttt{task\_NN}, part files become \texttt{part\_<i>}, and ground-truth links and joints become \texttt{link\_<i>} / \texttt{joint\_<i>}, with the original mapping sealed in a file that only evaluation scripts read.
Materials are replaced by one flat color per link, and the same colors appear in the meshes, in the template, and in the reference image.

\paragraph{Part meshes.}
Every part is one OBJ mesh that holds the merged geometry of one link, expressed in the frame of that link, whose origin is the pivot of the joint above it.
Every non-root part is rotated by a random rotation about the link origin (uniform on $SO(3)$, seeded), and uniformly scaled so that its longest bounding-box side is one.
The root part is given upright at its true size, which fixes the world frame and the metric scale of the object.
The world frame has $+Z$ up and $-Y$ as the front of the object, and lengths are in metres.

\paragraph{URDF template.}
The template gives the link tree, the link-to-part mapping, the joint types, the joint limits, and the mimic couplings.
Four kinds of fields are blanked with three question marks each, namely the visual rotation and the mesh scale of every non-root link and the origin and the axis of every joint, and the Executor recovers them from the part geometry and the reference image.
The frames of the links are never rotated, so a joint origin is the position of the child frame in the parent frame and a joint axis is a direction in world coordinates.
The positive direction of an axis is the direction of motion from the lower towards the upper limit, and a revolute pivot may be any point on the hinge line.
\hyperref[lst:template]{Listing~1} shows the template of an object with one revolute joint, including the header comment that states these conventions to the agent. Long lines of the listings are wrapped.

The output is the template with every blank filled, or an equivalent URDF exported by the SDK.
The benchmark also provides a view that keeps the true orientation and size of every part, and this paper uses the view described here, which we call rotated-and-scaled.

\phantomsection\label{lst:template}
\suppbox{lstgray}{Listing 1. URDF template of a test object (two links, one revolute joint)}{sec/listings/template_example.tex}

\section{Pipeline Details}
\label{sec:supp:pipeline}

\subsection{Executor}
\label{sec:supp:executor}

\paragraph{Prompt.}
The system prompt of the Executor is the Skill, i.e.\ the Markdown procedure followed by the contract of the Assembly SDK and the documentation of the helper scripts.
The user prompt is the same for every object and every Skill and only names the files of the task.
\suppbox{lstblue}{User prompt of the Executor}{sec/listings/executor_prompt.tex}
The Executor receives no description of the object, no category name, and no feedback from the GT-based Evaluator.
A session ends when the Executor stops calling tools or after $60$ tool turns.

\paragraph{Tools.}
The Executor's only tool is a persistent Python session (\texttt{execute\_\allowbreak python\_\allowbreak code}) into which the engine's SDK and a frozen \emph{Assembly SDK} are preloaded, plus read access to the files of the task and of the Skill.
Variables, imports, and helper functions survive across calls, and the paths of the task, the output, and the Skill are predefined as variables.
The helper scripts of the Skill are imported into the same session.

\paragraph{Assembly SDK.}
The Assembly SDK is the contract between the Skill and the frozen runtime.
Its twelve methods import and inspect the parts, set the scale and the visual pose of a part, create links and joints, set a joint from its world pivot and axis, move joints within their limits by forward kinematics, audit the tree, render previews and joint sweeps, and export the URDF.
Every call returns a JSON-safe value or raises a structured error with a stable error code.
The contract file is part of the Skill directory but read-only for the Skill Evolver, so every version of the Skill talks to the same runtime.
The SDK never places a part on its own, so a wrong rest pose is always a decision of the Executor.

\paragraph{Runtime Validator.}
The Runtime Validator of \secref{method:skill} consists of the \texttt{audit} method of the SDK and of the helper scripts that the Skill ships.
The audit checks that the tree is complete and that the source vertices of every part are unchanged.
Previews and joint sweeps, with four positions from two cameras, are rendered inside the rollout and inspected by the Executor.
The Skill Evolver can revise the checks that the Skill defines, while the audit of the SDK remains fixed.
Neither script-based role makes LLM calls.

\subsection{Seed and Improved Skills}
\label{sec:supp:procedure}

\paragraph{Seed Skill $S_0$.}
\hyperref[lst:seed]{Listing~2} reproduces the Markdown procedure of the seed Skill $S_0$ in full.
It states the input, the output, and the tools, and it contains no instruction on how to assemble an object, so every procedure in the final Skill was written by the Skill Evolver.
The seed Skill ships no helper script.

\paragraph{Final champion $S_9$.}
The champion $S_9$ keeps the text of $S_0$ unchanged and appends the five procedures of \hyperref[lst:s9]{Listing~3}, which grow the Markdown from $2.3$k to $9.3$k characters.
Each of the three accepted edits of Table~\ref{tab:lineage} contributed a part of them.
\begin{itemize}[leftmargin=*,itemsep=1pt,topsep=2pt]
    \item \emph{Root-registered rest-pose check} ($S_3$). Before the export, the Executor renders the complete assembly and compares it with the reference image through the helper \texttt{image\_compare.py}. The helper segments both images by the flat link colors, registers the render on the fixed root link, and returns the projected bounds and the center offset of every link. A displaced link of the right size and orientation then points to a misplaced pivot.
    \item \emph{Depth before screen-space corrections} ($S_6$). An image offset fixes two of the three coordinates of a pivot. The Skill asks the Executor to take the unseen coordinate from the attachment geometry of the parent, such as a bearing, a hinge edge, or a socket, and to hold this constraint while it corrects the projected coordinates.
    \item \emph{Guide engagement of one-sided sliders} ($S_6$). For a prismatic joint whose range starts at zero, the Executor evaluates both signs of the axis at the nonzero limit and keeps the sign under which the moving part stays engaged with its guide.
    \item \emph{Revolute bearings from circular rims} ($S_9$). The helper \texttt{round\_features.py} extracts closed circular sharp-edge loops from a mesh and returns their centers, normals, and radii. Rims of the child whose axis line passes near the mesh origin give candidate hinge axes, and matching them with the rims of the parent constrains both the rotation of the part and the joint axis.
    \item \emph{Transverse seating of sliders} ($S_9$). The two coordinates of a prismatic origin across the direction of travel are fixed by the rail, channel, or bore of the parent, and only the extension along the guide is fitted to the image.
\end{itemize}
Both helper scripts are deterministic, take meshes or images as input, and never read ground truth.
A third script is the preflight test that exercises every helper on synthetic input, and the contract gate runs it before a candidate is adopted.
All five procedures are stated in terms of generic mechanical features such as bearings, guides, and hinge edges.

\phantomsection\label{lst:seed}
\suppbox{lstgreen}{Listing 2. Seed Skill $S_0$ (Markdown procedure)}{sec/listings/skill_seed.tex}

\phantomsection\label{lst:s9}
\suppbox{lstgreen}{Listing 3. Text appended by the three accepted edits ($S_9$ = Listing 2 + Listing 3)}{sec/listings/skill_s9_added.tex}

\subsection{Skill Evolver}
\label{sec:supp:evolver}

\paragraph{System prompt.}
\hyperref[lst:evolver]{Listing~4} reproduces the parts of the system prompt of the Skill Evolver, which the code base calls the Skill Generator, that govern the search.
They state the goal, the bound of one small edit per failure cluster, the trial that every edit has to pass, the reading of the scores, and the ban on values taken from one object.
The omitted parts specify the files of a Skill, the documentation format of the helper scripts, the preflight test, and the journal.

\paragraph{Per-round prompt.}
The prompt of a round contains the task definition, the outcome of the previous candidate at the validation gate, the path of the journal of the previous round, the pass counts of every check on the train split for the recent rounds, and the failure clusters of the current round.
Every failed joint is one item, and the items are grouped by failure mode, joint type, and error size.
A cluster is listed with all of its member objects, the measured errors of every item, and the output directory of every session, as in the following abridged excerpt of round $3$.
\suppbox{lstorange}{Failure clusters in the prompt of round 3 (abridged)}{sec/listings/cluster_excerpt.tex}
Round $3$ of the reported run had $26$ failed objects in $23$ clusters, and the Skill Evolver works on the largest ones.

\paragraph{Tools.}
The Skill Evolver reads and edits files, searches the rollout artifacts, and runs code in its own simulation session.
Two further tools implement the first layer of the regression strategy and the contract gate.
\texttt{try\_skill} reruns the trial objects of one cluster with the edited Skill, scores them with the GT-based Evaluator, and returns the fitness before and after the edit.
\texttt{check\_skill} verifies that the SDK contract is unchanged, that every helper compiles and imports, that its documentation matches its signature, and that the preflight test passes.
A candidate that fails this check is handed back for repair up to three times.

\paragraph{Journal.}
At the end of its turn the Skill Evolver writes a journal of at most $300$ words on what it observed, investigated, changed, and verified, and on what remains unresolved.
The Evolver of the next round reads this journal first, which carries the reasoning across rounds without sharing a context.

\phantomsection\label{lst:evolver}
\suppbox{lstblue}{Listing 4. System prompt of the Skill Evolver (excerpt)}{sec/listings/evolver_system_prompt_excerpt.tex}

\subsection{Self-Improvement Loop}
\label{sec:supp:loop}

\begin{algorithm}[t]
\caption{Self-Improvement Loop with two-layer regression}
\label{alg:loop}
\small
\begin{algorithmic}[1]
\Require seed Skill $S_0$, train split $\mathcal{D}_{\mathrm{tr}}$, validation split $\mathcal{D}_{\mathrm{val}}$, rounds $T$, clusters $N_c$, objects $N_o$, attempts $M$, sessions $K$, threshold $z$
\Ensure champion Skill $S^\star$
\State $S^\star \gets S_0$
\State $\mathcal{B} \gets \textsc{Rollout}(S_0, \mathcal{D}_{\mathrm{tr}})$ \Comment{URDFs, artifacts, $\mathcal{E}$}
\State $F^\star \gets \textsc{Fitness}(S_0, \mathcal{D}_{\mathrm{val}}, K)$
\For{$t = 1, \dots, T-1$}
    \State $\mathcal{C} \gets \textsc{Cluster}(\mathcal{B})$ \Comment{recurring failures}
    \State $S_t \gets S^\star$ \Comment{fork from the champion}
    \For{each of the $N_c$ largest clusters $c \in \mathcal{C}$} \Comment{layer 1}
        \State $\mathcal{D}_c \gets$ $N_o$ train objects of $c$
        \For{$m = 1, \dots, M$}
            \State $S' \gets \textsc{Edit}(S_t, c)$ \Comment{Skill Evolver}
            \If{$\textsc{Fitness}(S', \mathcal{D}_c) > \textsc{Fitness}(S_t, \mathcal{D}_c)$}
                \State $S_t \gets S'$, \ \textbf{break}
            \EndIf
        \EndFor
    \EndFor
    \State $\mathcal{B} \gets \textsc{Rollout}(S_t, \mathcal{D}_{\mathrm{tr}})$
    \State $F_t \gets \textsc{Fitness}(S_t, \mathcal{D}_{\mathrm{val}}, K)$ \Comment{layer 2}
    \State $F^\star \gets F^\star \cup \textsc{Fitness}(S^\star, \mathcal{D}_{\mathrm{val}}, K)$ \Comment{pooled}
    \State $\Delta \gets$ mean per-object difference of $F_t$ and $F^\star$
    \If{$\Delta \geq z \cdot \mathrm{SE}(\Delta)$}
        \State $S^\star \gets S_t$, \ $F^\star \gets F_t$
    \EndIf
\EndFor
\State \Return $S^\star$
\end{algorithmic}
\end{algorithm}

Algorithm~\ref{alg:loop} gives the Self-Improvement Loop, and the rest of this subsection states what the pseudocode leaves out.
Ground-truth files and unredacted Evaluator outputs are held outside the Skill Evolver's file scope, and an audit of every agent transcript for ground-truth paths runs after each round.
The validation gate uses $K\,{=}\,3$ sessions per object and $z = 1.5$, and the standard error is that of the $40$ per-object differences in mean fitness.
The champion is run again in every round, and its sessions are pooled over rounds.
A rejected candidate whose difference is positive with $z \geq 0.5$ receives three more sessions per object, as does the champion, and the decision is repeated on the pooled sessions.
This happened once in the run, for $S_5$, which remained rejected.
After a rejection the next candidate is forked from the champion. The Skill Evolver is told of the rejection and still receives the rejected candidate's train rollouts and journal.
Executor code errors are retried once and transient provider or renderer errors up to six times with exponential back-off.
The run started from the hand-written seed Skill $S_0$ and received no human comments.

\subsection{Simulation and Rendering Engine}
\label{sec:supp:engine}
This subsection details the engine \emph{Light-3js-phyx} summarized in \secref{exp:setup}.

\begin{figure}[!htb]
    \centering
    \includegraphics[width=0.55\linewidth]{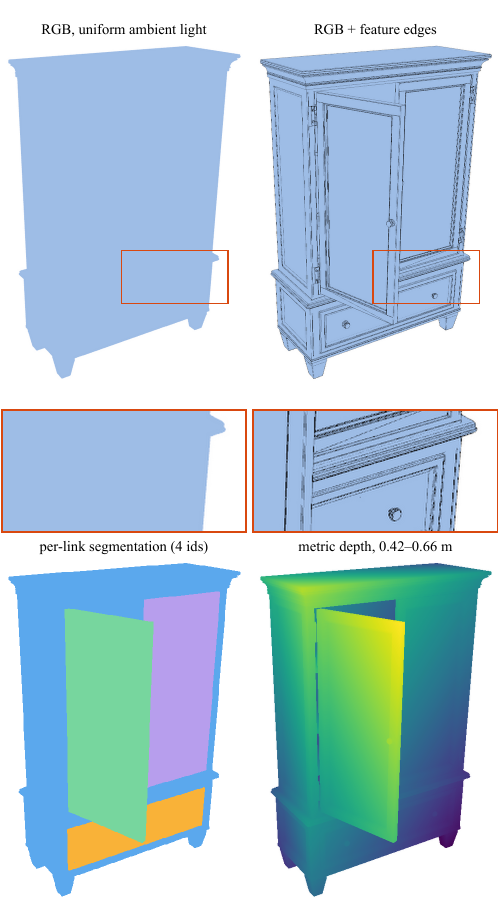}
    \caption{One capture, four aligned passes.
    A cabinet with two doors and a drawer is rendered from a single camera with one albedo on every link.
    The four passes are RGB under uniform ambient light, the same RGB with feature edges, a per-link segmentation, and metric depth.
    The insets enlarge the region around the closed drawer, whose seams are visible only with the feature edges.}
    \label{fig:engine}
\end{figure}

\paragraph{Interface and kinematics.}
The engine renders with three.js and simulates with Rapier in a headless Chromium.
The agents reach it through the Python session of \secref{supp:executor}, whose SDK loads a URDF, sets joint positions, places cameras, and captures images.
Every link is a kinematic body whose pose is written from forward kinematics, so the previews and the joint sweeps follow the URDF exactly.
The physics world reports contacts between bodies, and the checks of the GT-based Evaluator are computed from forward kinematics alone.

\paragraph{Rendering and throughput.}
Each capture yields up to three passes from one framebuffer, namely RGB, per-link segmentation, and metric depth, rendered without anti-aliasing so that they correspond pixel for pixel, as shown in \figref{engine}.
Lighting is uniform ambient, so a pixel shows the albedo of its material, and the RGB pass draws feature edges on top, which keep seams, panel gaps, and hinge sides visible at $512^2$.
The reference image of a task and the renders of the Executor use the same settings.
In the reported run, a pool of eight browsers with three environments each serves $14$ parallel Executor sessions on one $20$-core workstation.
On a $12$-core laptop, a pool of $50$ environments sustains about $1{,}800$ pass-frames per second at $640{\times}480$.

\subsection{Configuration}
\label{sec:supp:impl}
Table~\ref{tab:implementation} lists the configuration of the reported run.

\begin{table}[!htb]
    \centering
    \caption{Configuration of AgentAssemble in the reported run.
    $K$ is the number of independent Executor sessions per object, and $d$ is the diagonal of the ground-truth object.}
    \label{tab:implementation}
    \tablestyle{1.5pt}{1.1}
    \footnotesize
    \begin{tabular}{@{}ll@{}}
    \shline
    Parameter & Value \\
    \hline
    \multicolumn{2}{l}{\emph{Backbone (Executor and Skill Evolver)}} \\
    model / API & \texttt{gpt-6.1-sol} / OpenAI Responses \\
    control & reasoning effort (no sampling temperature) \\
    \hline
    \multicolumn{2}{l}{\emph{Executor}} \\
    reasoning effort / tool turns & low / $\leq 60$ \\
    tool & one MCP tool, persistent Python session \\
    runtime & frozen Assembly SDK (preloaded) \\
    sessions per object $K$ & $1$ train / $3$ validation / $3$ test \\
    retries & code error $1\times$, transient $6\times$ \\
    \hline
    \multicolumn{2}{l}{\emph{Runtime Validator (script)}} \\
    runtime audit & SDK structural audit \\
    Skill helpers & image comparison, circular-rim extraction \\
    renders per rollout & previews and sweeps, $4$ poses, $2$ cameras \\
    \hline
    \multicolumn{2}{l}{\emph{Skill Evolver}} \\
    reasoning effort & highest \\
    editable files & Skill Markdown, helpers, subagent prompts \\
    inputs & clustered $\mathcal{E}$, artifacts (no GT) \\
    own trials & each edit tested on $3$ train objects \\
    \hline
    \multicolumn{2}{l}{\emph{Self-Improvement Loop}} \\
    seed Skill $S_0$ & hand-written (\texttt{v\_io\_sdk\_rs}) \\
    batch / rounds & full train split ($40$) / $12$ \\
    fitness & mean graded score $\phi$, Eq.~\eqref{eq:fitness} \\
    candidate gate & contract (compile, import, smoke test) \\
    validation gate & paired, $40 \times 3$, $z \geq 1.5$ \\
    contract-repair attempts & $\leq 3$ \\
    human comments & none \\
    \hline
    \multicolumn{2}{l}{\emph{GT-based Evaluator (script)}} \\
    checks & $12$ scored, see Table~\ref{tab:oracle_checks} \\
    axis / limit tolerance & $10^\circ$ / $15\%$ rel.\ or $0.05$ abs. \\
    pose tolerance $\tau$ & $\max(2\,\text{cm}, 0.04\,d)$ \\
    sweep samples per joint & $5$ \\
    \hline
    \multicolumn{2}{l}{\emph{Simulation and hardware}} \\
    engine & three.js + Rapier, headless Chromium \\
    parallel Executor sessions & $14$ in the run, $20$ in test evaluations \\
    render resolution & $512^2$ \\
    workstation & $20$-core Linux, 46\,GB \\
    \hline
    \multicolumn{2}{l}{\emph{Data and evaluation}} \\
    dataset / view & ArtiAssembly subset / rotated-and-scaled \\
    splits & $40$ / $40$ / $40$, see Table~\ref{tab:splits} \\
    evaluation & same Executor config, Evolver disabled \\
    uncertainty & paired difference, standard error \\
    \shline
    \end{tabular}
\end{table}

\section{Dataset Details}
\label{sec:supp:benchmark}
\label{sec:supp:dataset}

\begin{figure}[!htb]
    \centering
    \includegraphics[width=0.62\linewidth]{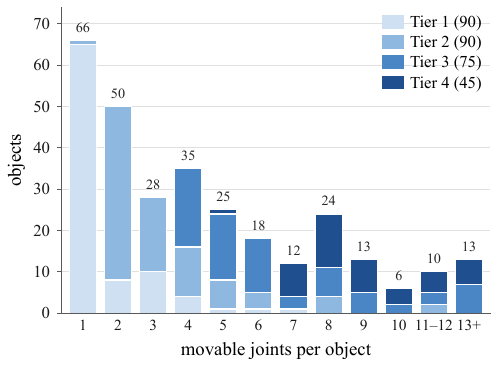}
    \caption{The $300$ objects of \emph{ArtiAssembly} by number of movable joints, split by tier.
    The legend gives the size of each tier.}
    \label{fig:dataset_stats}
\end{figure}

\paragraph{Composition.}
\emph{ArtiAssembly} has $300$ objects in four tiers of planned complexity, T1 to T4, with $90$, $90$, $75$, and $45$ objects and a mean of $1.6$, $3.4$, $6.9$, and $9.6$ movable joints per object.
\figref{dataset_stats} shows the number of objects by tier and by number of movable joints.
The objects cover $26$ domains, of which the largest are furniture with $41$ objects, tools with $28$, appliances with $27$, and toys with $26$.

\paragraph{Evaluation subset.}
The seed Skill $S_0$ is run with three sessions on the $213$ eligible objects, which have at most eight joints and all links visible in the reference image.
It never solves $105$ of them and always solves $53$, and the remaining $55$ objects are left out.
The subset draws $96$ objects from the first group and $24$ from the second at random, and every split receives $32$ and $8$ of them.
Table~\ref{tab:splits} gives the composition of the three splits.

\begin{table}[!htb]
    \centering
    \caption{ArtiAssembly and the splits of the evaluation subset (rotated-and-scaled view).
    Dom.: number of domains.
    T1--T4: number of objects of each tier.
    Rev./Pris./Cont.: revolute/prismatic/continuous joint counts.}
    \label{tab:splits}
    \tablestyle{2.5pt}{1.1}
    \begin{tabular}{@{}lrrrrrrrrrr@{}}
    \shline
    Split & Inst. & Dom. & T1 & T2 & T3 & T4 & Links & Rev. & Pris. & Cont. \\
    \hline
    Train      & 40  & 14 & 12 & 20 & 6  & 2  & 172  & 69  & 30  & 32 \\
    Validation & 40  & 19 & 14 & 12 & 10 & 4  & 184  & 86  & 27  & 20 \\
    Test       & 40  & 17 & 14 & 15 & 7  & 4  & 166  & 63  & 34  & 28 \\
    \hline
    Subset     & 120 & 23 & 40 & 47 & 23 & 10 & 522  & 218 & 91  & 80 \\
    Eligible   & 213 & 25 & 84 & 75 & 36 & 18 & 885  & 368 & 160 & 129 \\
    ArtiAssembly   & 300 & 26 & 90 & 90 & 75 & 45 & 1723 & 670 & 290 & 440 \\
    \shline
    \end{tabular}
\end{table}

\paragraph{Difficulty labels.}
Independently of its tier, we label each instance by its number of movable joints as \emph{easy} (one joint), \emph{medium} (two or three), or \emph{hard} (four or more), and the comparison table of the main paper uses this label.

\paragraph{Duplicates.}
A per-instance signature, the sorted multiset of per-part vertex digests, confirms that no two objects of \emph{ArtiAssembly} have exactly the same part geometry.

\section{Oracle Details}
\label{sec:supp:oracle}

The system has two oracles.
The GT-based Evaluator scores an exported URDF against the ground truth and is described in this section, and the Runtime Validator checks an assembly without ground truth and is described in \secref{supp:executor}.

\begin{figure*}[!htb]
    \centering
    \includegraphics[width=\textwidth]{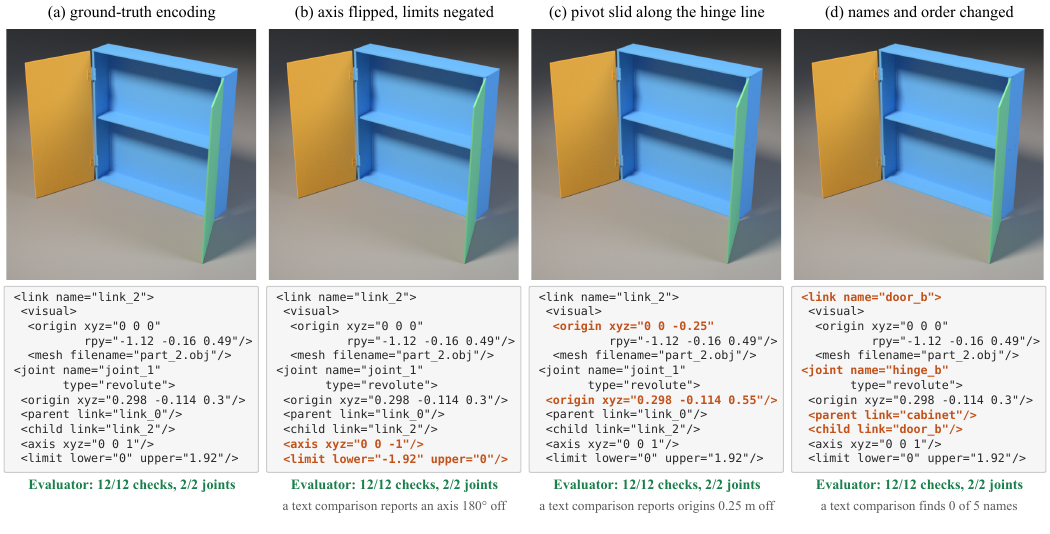}
    \caption{The same mechanism, four URDF encodings.
    A two-door cabinet is rendered with both doors at their joint limit under four URDFs that are textually different and physically identical. They are (a) the ground-truth encoding, (b) the hinge axes flipped and the limits negated, (c) the pivots moved by $0.25$\,m along the hinge lines with the opposite offset in the visual origins, and (d) links and joints renamed and listed in reverse order.
    Highlighted lines mark what changed relative to (a).
    A comparison of the text would report a $180^\circ$ axis error, a $0.25\,\text{m}$ origin error, or no matching names, whereas the GT-based Evaluator canonicalizes the encodings and matches links by geometry, so all four pass $12/12$ checks with both joints correct.}
    \label{fig:ambiguity}
\end{figure*}

\paragraph{Canonicalization and link matching.}
The same mechanism admits many URDF encodings, as \figref{ambiguity} shows.
The Evaluator therefore brings both URDFs into a canonical form before any comparison.
It applies the mesh scale and the visual transform to the geometry, expresses every joint axis in world coordinates at rest, and flips an axis together with its limits so that its dominant component is positive.
The links of the prediction are then matched to the ground-truth links by a signature of their mesh geometry that is invariant to rigid motion, with one globally optimal assignment.
Parts with identical geometry are interchangeable, and the assignment that scores best is kept.
All later checks compare matched links, so names and the order of elements have no influence.

\paragraph{Tolerance.}
The pose checks share one tolerance that scales with the object, $\tau = \max(2\,\text{cm},\, 0.04\,d)$, where $d$ is the diagonal of the ground-truth object.

\paragraph{Structural and articulation checks.}
Table~\ref{tab:oracle_checks} lists the checks, of which the structural ones and the counts are exact comparisons.
\emph{Topology} compares the edge sets of the two trees under the link matching.
\emph{Joint types} counts revolute and continuous as one rotational type.
\emph{Joint axes} requires the world-space axis of every movable joint to lie within $10^\circ$ of the ground truth, and \emph{joint limits} requires both limits to agree within $15\%$ relative or $0.05$ absolute tolerance.

\paragraph{Geometric checks.}
\emph{Rest pose} measures, for every matched link at $\mathbf{q} = \mathbf{0}$, the symmetric distance between the surface of the predicted link and the surface of the ground-truth link, and requires it to stay within $\tau$.
The distance is computed from points on one surface to the triangles of the other, in both directions, and the largest value is taken.
The points of a link are its vertices, up to $1{,}500$ of them, and $1{,}000$ samples drawn uniformly over its area with a fixed seed.
It is invariant to the tessellation and to any symmetry of the part, so a wheel or a knob that is mounted rotated about its own axis counts as correct, and a thin plate that is mounted upside down is caught.
\emph{Joint pivot} applies to revolute joints and requires the distance between the predicted and the ground-truth axis lines to be at most $\tau$. A shift along the axis describes the same hinge line and is ignored.
\emph{Kinematic equivalence} drives each movable joint through five evenly spaced positions from its lower to its upper limit while the other joints stay fixed, and requires the links that the joint moves to stay within $\tau$ of their ground-truth trajectory under the same surface distance.
A continuous joint is driven over one full turn, and a joint with a mimic coupling is driven on its own. Both parameter directions are evaluated and the better one is kept, so the sign convention of $q$ has no influence.
\emph{Link scale} compares the size of every matched mesh with the ground truth within $5\%$. It is reported as a diagnostic and is left out of the pass predicate.

\begin{figure*}[!htb]
    \centering
    \includegraphics[width=\textwidth]{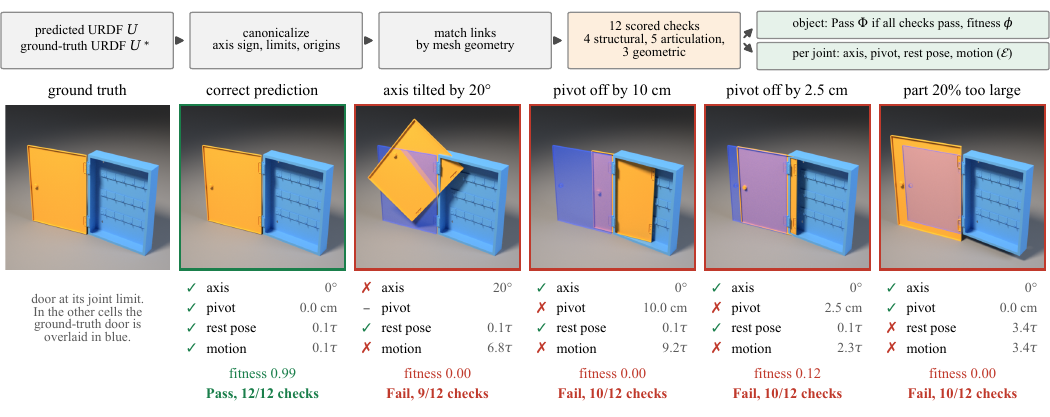}
    \caption{How assemblies are scored.
    \emph{Top}: the predicted and ground-truth URDFs are canonicalized, their links are matched by geometry, and the checks of Table~\ref{tab:oracle_checks} yield the object-level outputs and a per-joint record.
    \emph{Bottom}: a cabinet with one door, rendered with the door at its joint limit, under a correct prediction and four perturbations of the door. The ground-truth door is overlaid in translucent blue.
    Below each render we list the per-joint dimensions with the measured error, namely the axis angle, the distance of the axis lines, and the rest-pose and trajectory drift in units of the tolerance $\tau$, followed by the fitness and the verdict of the Evaluator.
    Each perturbation fails the dimension it targets together with the motion. A tilted axis leaves the pivot without a comparison (marked ``--''), a pivot that is off by slightly more than $\tau$ still earns partial fitness, and a part of the wrong size fails the rest pose while its axis and pivot are correct.}
    \label{fig:metrics}
\end{figure*}

\paragraph{Outputs.}
A session passes when all twelve scored checks pass.
For every movable joint of the ground truth the Evaluator records whether its axis, its pivot, and its trajectory are within tolerance, together with the axis angle, the distance of the axis lines, and the drift $\delta_j$ of the trajectory, from which the fitness of Eq.~\eqref{eq:fitness} follows.
For a link that fails the rest pose, the Evaluator also names the smallest correction that would bring it within tolerance. It tries a change of size, of orientation about the pivot, and of the pivot position, first alone and then in pairs.
These records are the diagnostics $\mathcal{E}$ from which the failure clusters of \secref{supp:evolver} are formed.
\figref{metrics} illustrates the scoring on one object with four perturbed hinges.

\paragraph{Role of the checks in this task.}
The template fixes the tree, the joint types, and the limits, so the structural checks and three of the articulation checks pass in every session that exports a complete URDF.
The discrimination comes from the joint axes and from the three scored geometric checks.

\begin{table}[!htb]
    \centering
    \caption{The checks of the GT-based Evaluator.
    A session passes if every scored check passes.
    $\tau = \max(2\,\text{cm}, 0.04\,d)$, with $d$ the diagonal of the ground-truth object.}
    \label{tab:oracle_checks}
    \tablestyle{2.5pt}{1.1}
    \begin{tabular}{@{}lll@{}}
    \shline
    Check & Compares & Tolerance \\
    \hline
    \multicolumn{3}{@{}l}{\emph{Structural}} \\
    model integrity & meshes and transforms valid & exact \\
    valid tree & single connected tree & exact \\
    link count & number of links & exact \\
    link matching & GT links matched by geometry & $100\%$ \\
    \hline
    \multicolumn{3}{@{}l}{\emph{Articulation}} \\
    joint type counts & count of each joint type & exact \\
    topology & edge set under matching & exact \\
    joint types & type of each joint$^\dagger$ & exact \\
    joint axes & world-space axis direction & $\leq 10^\circ$ \\
    joint limits & lower and upper limit & $15\%$ / $0.05$ \\
    \hline
    \multicolumn{3}{@{}l}{\emph{Geometric}} \\
    link scale$^\ddagger$ & size ratio of each mesh & $\pm 5\%$ \\
    rest pose & link surfaces at $q\,{=}\,0$ & $\leq \tau$ \\
    joint pivot & revolute axis line & $\leq \tau$ \\
    kinematic equiv. & link trajectories over a sweep & $\leq \tau$ \\
    \shline
    \multicolumn{3}{@{}l}{\footnotesize $^\dagger$revolute and continuous count as one rotational type.} \\
    \multicolumn{3}{@{}l}{\footnotesize $^\ddagger$reported as a diagnostic and left out of the pass predicate.} \\
    \end{tabular}
\end{table}

\section{Additional Results}
\label{sec:supp:results}
Table~\ref{tab:lineage} lists every round of the run.
Six candidates were rejected with a difference from the champion between $-0.019$ and $+0.012$ fitness.
Several of them had improved the train objects on which the Skill Evolver tested them, for example $S_4$, whose train fitness of $0.587$ exceeded that of the champion $S_3$ by $0.09$ while its validation fitness was $0.013$ lower.

\begin{table*}[!htb]
    \centering
    \caption{Lineage of the run.
    Each row is the candidate Skill of one round and the edit the Skill Evolver made to produce it. Every candidate is forked from the champion held at that time.
    Train and validation columns are measured on the respective splits.
    The validation columns give the evaluation of each candidate at its gate.
    $\Delta$ is the mean per-object difference in validation fitness between the candidate and the champion, and $z$ is that difference in standard errors. The gate accepts at $z \geq 1.5$.}
    \label{tab:lineage}
    \tablestyle{4pt}{1.1}
    \begin{tabular}{@{}ly{185}ccccccl@{}}
    \shline
    Version & Edit by the Skill Evolver & \multicolumn{2}{c}{Train} & \multicolumn{2}{c}{Validation} & $\Delta$ & $z$ & Gate \\
     & & Fitness & Pass (\%) & Fitness & Pass (\%) & & & \\
    \hline
    $S_0$ & hand-written seed Skill $S_0$ & 0.431 & 25.0 & 0.439 & 24.2 & -- & -- & seed \\
    $S_1$ & no edit survived the Skill Evolver's own trials, Skill unchanged & 0.427 & 25.0 & -- & -- & -- & -- & unchanged \\
    $S_2$ & cavity-mouth detection for hollow parts from orthographic depth (helper \texttt{cavity.py}) & 0.448 & 27.5 & 0.455 & 29.2 & $-0.002$ & $-0.07$ & rejected \\
    $S_3$ & root-registered comparison of the render with the reference image, per-link projected bounds and offsets (helper \texttt{image\_compare.py}) & 0.496 & 35.0 & 0.544 & 37.5 & $+0.089$ & $+2.59$ & \textbf{accepted} \\
    $S_4$ & paired pivot-hardware axes from local rotational evidence (helper \texttt{axis\_evidence.py}), and prismatic sign from measured seats & 0.587 & 47.5 & 0.538 & 35.8 & $-0.013$ & $-0.45$ & rejected \\
    $S_5$ & all joint frames synchronised parent-first before the visual poses are set & 0.549 & 37.5 & 0.550 & 35.0 & $+0.001$ & $+0.03$ & rejected \\
    $S_6$ & attachment depth constrained by the parent geometry before screen-space corrections, and one-sided slider endpoints checked against guide engagement & 0.694 & 50.0 & 0.601 & 38.3 & $+0.061$ & $+2.78$ & \textbf{accepted} \\
    $S_7$ & scale of pivoted parts fitted from anchored distal landmarks & 0.664 & 52.5 & 0.584 & 38.3 & $-0.019$ & $-0.82$ & rejected \\
    $S_8$ & hinge-angle correction from directed edges in the reference image (helper \texttt{projected\_direction.py}) & 0.652 & 47.5 & 0.606 & 40.8 & $+0.012$ & $+0.47$ & rejected \\
    $S_9$ & closed circular rims extracted as bearing-axis and center evidence (helper \texttt{round\_features.py}), and slider placement constrained transversally before the extension is fitted & 0.714 & 60.0 & 0.682 & 52.5 & $+0.086$ & $+3.00$ & \textbf{accepted} \\
    $S_{10}$ & no edit survived the Skill Evolver's own trials, Skill unchanged & 0.716 & 57.5 & -- & -- & -- & -- & unchanged \\
    $S_{11}$ & signed joint motion checked numerically against the parent geometry, and a half-turn test on surface relief & 0.698 & 55.0 & 0.670 & 51.7 & $-0.016$ & $-0.77$ & rejected \\
    \shline
    \end{tabular}
\end{table*}

\subsection{Transfer to test objects}
\label{sec:supp:transfer}
Table~\ref{tab:groups} breaks the test result of the seed Skill $S_0$ and of the three accepted Skills down by group of objects.
The objects that the seed Skill $S_0$ never solved during selection account for the gain, as their Pass Rate rises from $14.6\%$ to $40.6\%$, and the objects it always solved stay solved under every Skill.
The gain appears at every difficulty label of the main paper.
The Pass Rate rises from $66.7\%$ to $84.8\%$ on easy objects, from $24.4\%$ to $53.3\%$ on medium objects, and from $7.1\%$ to $26.2\%$ on hard objects, which remain the hard part of the task.
The last two columns repeat the comparison with \texttt{gpt-6-astra} as the Executor.
The pattern is the same at a higher level. The never-solved objects rise from $29.2\%$ to $64.6\%$, medium objects from $44.4\%$ to $82.2\%$, and hard objects from $21.4\%$ to $40.5\%$.

\begin{table}[!htb]
    \centering
    \caption{Test Pass Rate (\%) and fitness of the seed Skill $S_0$ and the accepted Skills by group of objects.
    The first four columns use \texttt{gpt-6.1-sol} as the Executor and the last two use \texttt{gpt-6-astra}.
    Easy objects have one movable joint, medium objects two or three, and hard objects four or more.
    $n$ is the number of objects.}
    \label{tab:groups}
    \tablestyle{2.5pt}{1.1}
    \begin{tabular}{@{}lrcccc|cc@{}}
    \shline
     & & \multicolumn{4}{c|}{sol} & \multicolumn{2}{c}{astra} \\
    Group & $n$ & $S_0$ & $S_3$ & $S_6$ & $S_9$ & $S_0$ & $S_9$ \\
    \hline
    \multicolumn{8}{@{}l}{\emph{Pass Rate (\%)}} \\
    all objects & 40 & 30.0 & 34.2 & 45.8 & 52.5 & 43.3 & 71.7 \\
    never solved by $S_0$ & 32 & 14.6 & 18.8 & 34.4 & 40.6 & 29.2 & 64.6 \\
    always solved by $S_0$ & 8 & 91.7 & 95.8 & 91.7 & 100.0 & 100.0 & 100.0 \\
    easy & 11 & 66.7 & 66.7 & 84.8 & 84.8 & 69.7 & 97.0 \\
    medium & 15 & 24.4 & 37.8 & 42.2 & 53.3 & 44.4 & 82.2 \\
    hard & 14 & 7.1 & 4.8 & 19.0 & 26.2 & 21.4 & 40.5 \\
    \hline
    \multicolumn{8}{@{}l}{\emph{Fitness}} \\
    all objects & 40 & 0.474 & 0.512 & 0.639 & 0.681 & 0.568 & 0.814 \\
    never solved by $S_0$ & 32 & 0.373 & 0.414 & 0.578 & 0.612 & 0.472 & 0.775 \\
    always solved by $S_0$ & 8 & 0.880 & 0.903 & 0.881 & 0.956 & 0.952 & 0.971 \\
    easy & 11 & 0.644 & 0.675 & 0.838 & 0.842 & 0.690 & 0.933 \\
    medium & 15 & 0.425 & 0.488 & 0.626 & 0.669 & 0.531 & 0.876 \\
    hard & 14 & 0.393 & 0.409 & 0.496 & 0.567 & 0.513 & 0.653 \\
    \shline
    \end{tabular}
\end{table}

\subsection{Transfer across Executor models}
\label{sec:supp:models}
Table~\ref{tab:test} lists the values plotted in \figref{models} together with the paired comparison with the seed Skill $S_0$ for each Executor model on the test split.
Every accepted Skill improves on the seed Skill $S_0$ under every model, and the gain of $S_6$ and $S_9$ exceeds $2.6$ standard errors in all three.
Under \texttt{claude-opus-5-5}, Pass Rate stops rising after $S_6$ while fitness continues to rise.
We also ran $S_0$ and $S_9$ with \texttt{gpt-6-astra}, the model of the USDCraft comparison in \secref{exp:comparison}.
It is the strongest of the four Executors under both Skills, and $S_9$ raises its fitness by $0.246 \pm 0.049$ and its Pass Rate from $43.3\%$ to $71.7\%$.
The gain of $S_9$ is therefore of similar size under all four models, between $0.165$ and $0.256$ fitness, although the Skill was improved with \texttt{gpt-6.1-sol} only.

\begin{table}[!htb]
    \centering
    \caption{Test-split results of the seed Skill $S_0$ and the three accepted Skills under four Executor models.
    The Skills were obtained with \texttt{gpt-6.1-sol} only, and \texttt{gpt-6-astra} was run with $S_0$ and $S_9$.
    Fit.: fitness. Pass: Pass Rate in percent.
    $\Delta$ is the mean per-object difference in fitness from $S_0$ with its standard error, $z$ is $\Delta$ in standard errors, W/L counts the objects whose fitness rises or falls, and $\Delta$ Pass is in percentage points.}
    \label{tab:test}
    \tablestyle{1.6pt}{1.1}
    \begin{tabular}{@{}llcccccc@{}}
    \shline
    Executor & Skill & Fit. & Pass & $\Delta$ & $z$ & W/L & $\Delta$ Pass \\
    \hline
    sol & $S_0$ & 0.474 & 30.0 & -- & -- & -- & -- \\
     & $S_3$ & 0.512 & 34.2 & $+0.038 \pm 0.029$ & 1.3 & 24/15 & $+4.2 \pm 5.0$ \\
     & $S_6$ & 0.639 & 45.8 & $+0.165 \pm 0.036$ & 4.6 & 34/6 & $+15.8 \pm 5.2$ \\
     & $S_9$ & 0.681 & 52.5 & $+0.207 \pm 0.041$ & 5.1 & 32/8 & $+22.5 \pm 5.5$ \\
    \hline
    opus & $S_0$ & 0.387 & 18.3 & -- & -- & -- & -- \\
     & $S_3$ & 0.471 & 30.8 & $+0.084 \pm 0.042$ & 2.0 & 24/15 & $+12.5 \pm 5.2$ \\
     & $S_6$ & 0.496 & 33.3 & $+0.109 \pm 0.041$ & 2.6 & 30/10 & $+15.0 \pm 5.2$ \\
     & $S_9$ & 0.552 & 33.3 & $+0.165 \pm 0.034$ & 4.8 & 32/7 & $+15.0 \pm 4.8$ \\
    \hline
    terra & $S_0$ & 0.316 & 18.3 & -- & -- & -- & -- \\
     & $S_3$ & 0.415 & 29.2 & $+0.099 \pm 0.042$ & 2.4 & 25/12 & $+10.8 \pm 5.4$ \\
     & $S_6$ & 0.534 & 38.3 & $+0.218 \pm 0.044$ & 5.0 & 36/4 & $+20.0 \pm 5.3$ \\
     & $S_9$ & 0.572 & 39.2 & $+0.256 \pm 0.039$ & 6.5 & 35/5 & $+20.8 \pm 4.7$ \\
    \hline
    astra & $S_0$ & 0.568 & 43.3 & -- & -- & -- & -- \\
     & $S_9$ & 0.814 & 71.7 & $+0.246 \pm 0.049$ & 5.1 & 33/7 & $+28.3 \pm 7.0$ \\
    \shline
    \end{tabular}
\end{table}

\subsection{What improves and what does not}
\label{sec:supp:checks}
Table~\ref{tab:checks} gives the share of test sessions that pass each discriminating check.
The structural checks, the joint types, and the limits pass in every session that exports a URDF.
The rest pose and the pivot gain the most under both Executor models, and the rest pose remains the check that fails most often.
At the level of joints with \texttt{gpt-6.1-sol}, pivot lines within tolerance rise from $83.6\%$ to $94.4\%$ and joint trajectories from $39.0\%$ to $55.1\%$, and the median trajectory drift falls from $1.54\,\tau$ to $0.81\,\tau$.
With \texttt{gpt-6-astra} the median drift falls from $1.05\,\tau$ to $0.49\,\tau$.

\begin{table}[!htb]
    \centering
    \caption{Share of test sessions (\%) that pass each check under the seed Skill $S_0$ and under $S_9$.}
    \label{tab:checks}
    \tablestyle{5pt}{1.1}
    \begin{tabular}{@{}lcc|cc@{}}
    \shline
     & \multicolumn{2}{c|}{sol} & \multicolumn{2}{c}{astra} \\
    Check & $S_0$ & $S_9$ & $S_0$ & $S_9$ \\
    \hline
    joint axes & 81.7 & 90.8 & 85.8 & 94.2 \\
    rest pose & 35.8 & 59.2 & 53.3 & 80.0 \\
    joint pivot & 71.7 & 87.5 & 78.3 & 93.3 \\
    kinematic equiv. & 30.0 & 53.3 & 43.3 & 71.7 \\
    \shline
    \end{tabular}
\end{table}

\subsection{Qualitative results under other Executor models}
\label{sec:supp:qualitative}
\begin{figure*}[!htb]
    \centering
    \includegraphics[width=\textwidth]{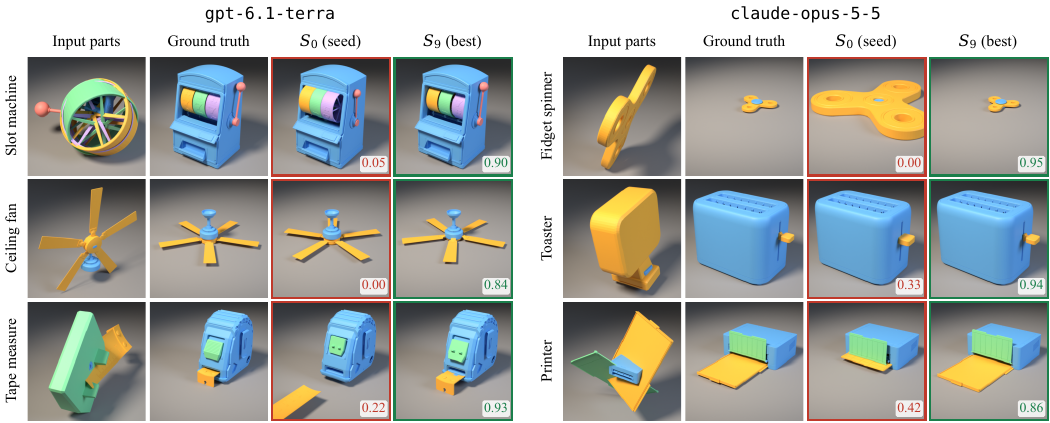}
    \caption{Assemblies under the seed Skill $S_0$ and under $S_9$ with the two Executor models that took no part in the self-improvement, in the layout of \figref{qualitative}.
    For each model, the objects are chosen among the test objects on which the seed Skill $S_0$ fails and $S_9$ passes.}
    \label{fig:qualitative_models}
\end{figure*}

\figref{qualitative_models} shows assemblies under the seed Skill $S_0$ and under $S_9$ with \texttt{gpt-6.1-terra} and \texttt{claude-opus-5-5}, the two Executor models that took no part in the self-improvement.

\subsection{Comparison with existing methods}
\label{sec:supp:comparison}
This section gives the protocol behind Table~\ref{tab:comparison} and the metrics that the main paper omits.

\noindent\textbf{Inputs of the two methods.}
Particulate~\cite{liParticulateFeedForward3D2026} and USDCraft~\cite{zhangUSDCraftGeometricallyGrounded2026} articulate one static mesh.
For every test object we therefore pose the ground-truth URDF at rest and merge all links into a single mesh without part labels, materials, or groups.
USDCraft receives this mesh and the reference image of the task, and we run it with \texttt{gpt-6-astra} at low reasoning effort and a budget of $80$ turns per object.
It returned a URDF for all $40$ test objects.
Particulate receives the same mesh with $+Z$ as the up direction and samples $32{,}768$ surface points, the largest number that fits the $24$\,GB of our GPU.
Our Executor receives the separated parts at a random rotation and unit scale together with the template, as in every other experiment.

\noindent\textbf{Scoring.}
Both methods output their own geometry, so the links of a prediction cannot be identified by their meshes.
We sample $3{,}000$ surface points on every predicted and ground-truth link at the rest pose and match the two sets of links with the Hungarian algorithm on the symmetric chamfer distance.
Every movable joint of the ground truth is then scored against the joint above its matched link, on the type, the axis within $10^\circ$, the distance between the two axis lines for revolute joints, and the drift of the ground-truth subtree when it is swept by the predicted joint over the ground-truth range, under the better of the two sign conventions.
The tolerance $\tau$ is that of the GT-based Evaluator.
Joint limits are left unscored, because our template states them.
The test split has $125$ movable joints, of which $91$ are revolute.
Our method is scored on each of its three sessions per object with the same script, and we report the mean, and the rows of the two existing methods come from one prediction per object.

\begin{table}[!htb]
    \centering
    \caption{Joint-level comparison on the ArtiAssembly test split with the structural columns and two further Executor models.
    Links: objects with the correct number of links.
    Topo.: objects whose matched links all hang from the correct parent.
    Type: share of the $125$ ground-truth movable joints with the correct type.
    Joint and Obj.\ are the columns of Table~\ref{tab:comparison} over all objects.
    All values are percentages.}
    \label{tab:supp_comparison}
    \tablestyle{3pt}{1.1}
    \begin{tabular}{@{}llccccc@{}}
    \shline
    Method & Model & Links & Topo. & Type & Joint & Obj. \\
    \hline
    Particulate & -- & 27.5 & 30.0 & 35.2 & 19.2 & 22.5 \\
    USDCraft & astra & 60.0 & 82.5 & 87.2 & 83.2 & 75.0 \\
    \hline
    Ours, $S_0$ & sol & 99.2 & 98.3 & 99.7 & 76.0 & 62.5 \\
    Ours, $S_9$ & sol & 99.2 & 99.2 & 99.2 & 89.1 & 85.8 \\
    Ours, $S_0$ & astra & 100.0 & 100.0 & 100.0 & 81.6 & 70.0 \\
    Ours, $S_9$ & astra & 100.0 & 99.2 & 99.5 & \textbf{93.6} & \textbf{92.5} \\
    Ours, $S_9$ & opus & 100.0 & 97.5 & 98.4 & 78.7 & 64.2 \\
    Ours, $S_9$ & terra & 100.0 & 99.2 & 98.9 & 81.6 & 70.8 \\
    \shline
    \end{tabular}
\end{table}

\noindent\textbf{Further results.}
Table~\ref{tab:supp_comparison} extends the comparison.
The structural columns show what the difference in inputs means.
USDCraft has to segment the mesh and recovers the correct number of links on $60.0\%$ of the objects and the correct tree on $82.5\%$, whereas our template states both. The few missing points in our rows are sessions that exported no URDF or in which a misplaced part was matched to another ground-truth link.
The last two rows score $S_9$ under the two Executor models of \secref{exp:models} with the same script.
With \texttt{claude-opus-5-5} and \texttt{gpt-6.1-terra}, $S_9$ recovers $78.7\%$ and $81.6\%$ of the joints, against $83.2\%$ for USDCraft, and $64.2\%$ and $70.8\%$ of the objects, against $75.0\%$.
The lead over USDCraft in Table~\ref{tab:comparison} therefore holds for the two stronger Executor models.

\subsection{Ablation runs}
\label{sec:supp:ablation}
This section details the variants of Table~\ref{tab:ablation}.
Each variant is one run of twelve rounds from the seed Skill $S_0$ with the configuration of Table~\ref{tab:implementation}, and it differs from the full method only in what is stated below.
Table~\ref{tab:abl_test} gives the test result of the Skills that we evaluated.

\noindent\textbf{Without regression strategy.}
Both layers are removed.
The Skill Evolver does not test its edits on train objects, no candidate is evaluated on the validation split, and the Skill that the Evolver writes in round $t$ is used in round $t{+}1$ whatever its result.
The Evolver edited the Skill in every round, with two or three edits per round.
This variant has no champion, so Table~\ref{tab:ablation} reports its Skill of round $9$, the round in which the full method wrote its final champion, and Table~\ref{tab:abl_test} adds the Skills of rounds $3$ and $6$.
The train fitness reaches $0.594$ in round $3$ and $0.605$ in round $7$ and then falls to $0.524$ in the last round, whereas the full method ends at $0.698$.
The same course appears on the test split. The Skills of rounds $3$ and $6$ improve over $S_0$ by $0.09$ fitness, and the Skill of round $9$ is back at the level of $S_0$.
Early edits thus help in this variant too, and later edits undo the gain because nothing removes an edit that hurts.
The Skill also grows without bound, to $35.7$k characters of Markdown and four helper scripts in round $9$, against $9.3$k characters and two helper scripts for $S_9$ of the full method.

\noindent\textbf{Without diagnostics.}
The Skill Evolver receives the pass predicate and the fitness of every train rollout and nothing else.
The per-check results and the per-joint records are withheld, and the Evaluator's output files are moved out of its reach before every turn.
Without the diagnostics the failures cannot be grouped into clusters, so the Evolver makes one edit per round and tries it on one train object.
The validation gate is unchanged.
The gate accepted five of the eleven candidates, in rounds $1$, $4$, $6$, $7$, and $8$, with paired validation gains between $0.045$ and $0.098$ fitness.
On the test split the first two champions lie slightly below $S_0$, and the gain arrives with the champions of rounds $7$ and $8$, whose edits add an image comparison registered on the root and a closure check for members with two bearings.
The final champion reaches a fitness of $0.609$ and a Pass Rate of $47.5\%$, below the $0.681$ and $52.5\%$ of the full method.
With the score alone, the Evolver arrives at related procedures later in the run.

\begin{table}[!htb]
    \centering
    \caption{Test results of the Skills of the ablation runs and of the human-curated Skill, with \texttt{gpt-6.1-sol} as the Executor.
    The subscript is the round in which the Skill was written.
    $\Delta$ is the paired fitness difference from the seed Skill $S_0$ with its standard error, $z$ is $\Delta$ in standard errors, and W/L counts the objects whose fitness rises or falls.}
    \label{tab:abl_test}
    \tablestyle{2.5pt}{1.1}
    \begin{tabular}{@{}lccccc@{}}
    \shline
    Skill & Fitness & Pass (\%) & $\Delta$ & $z$ & W/L \\
    \hline
    Seed Skill $S_0$ & 0.474 & 30.0 & -- & -- & -- \\
    Human-curated Skill & 0.541 & 37.5 & $+0.067 \pm 0.024$ & 2.8 & 27/11 \\
    \hline
    \multicolumn{6}{@{}l}{\emph{w/o regression strategy}} \\
    $S_3$ & 0.566 & 40.0 & $+0.092 \pm 0.021$ & 4.4 & 32/8 \\
    $S_6$ & 0.563 & 40.8 & $+0.089 \pm 0.029$ & 3.1 & 28/11 \\
    $S_9$ & 0.475 & 30.0 & $+0.001 \pm 0.032$ & 0.0 & 18/22 \\
    \hline
    \multicolumn{6}{@{}l}{\emph{w/o diagnostics}, champions} \\
    $S_1$ & 0.437 & 28.3 & $-0.037 \pm 0.030$ & $-1.2$ & 20/19 \\
    $S_4$ & 0.421 & 25.0 & $-0.053 \pm 0.036$ & $-1.5$ & 20/19 \\
    $S_6$ & 0.504 & 34.2 & $+0.030 \pm 0.029$ & 1.1 & 27/13 \\
    $S_7$ & 0.596 & 44.2 & $+0.122 \pm 0.037$ & 3.3 & 31/8 \\
    $S_8$ & 0.609 & 47.5 & $+0.135 \pm 0.030$ & 4.5 & 33/7 \\
    \hline
    \multicolumn{6}{@{}l}{\emph{Full method}, champions} \\
    $S_3$ & 0.512 & 34.2 & $+0.038 \pm 0.029$ & 1.3 & 24/15 \\
    $S_6$ & 0.639 & 45.8 & $+0.165 \pm 0.036$ & 4.6 & 34/6 \\
    $S_9$ & \textbf{0.681} & \textbf{52.5} & $\mathbf{+0.207 \pm 0.041}$ & 5.1 & 32/8 \\
    \shline
    \end{tabular}
\end{table}

\noindent\textbf{Human-curated Skill.}
This Skill keeps the task statement of $S_0$ and adds a written procedure of $12.3$k characters without helper scripts.
It raises the fitness of $S_0$ by $0.067 \pm 0.024$ and the Pass Rate from $30.0\%$ to $37.5\%$.
Rest-pose alignment passes in $48.3\%$ of its test sessions, between the $35.8\%$ of $S_0$ and the $59.2\%$ of $S_9$, and pivot alignment in $72.5\%$, close to the $71.7\%$ of $S_0$ and below the $87.5\%$ of $S_9$.
A general procedure thus helps the placement of the parts, and the pivots improve with the measured attachment constraints that the Skill Evolver derived from the failures.

\end{document}